%% file: acl2023.tex
\pdfoutput=1

\documentclass[11pt]{article}

\usepackage[final]{ACL2023}

\usepackage{float}

\usepackage{times}
\usepackage{latexsym}

\usepackage[T1]{fontenc}

\usepackage[utf8]{inputenc}

\usepackage{microtype}

\usepackage{inconsolata}
\usepackage[dvipsnames]{xcolor}
\usepackage{amsmath}
\usepackage{multirow}
\usepackage{booktabs}
\usepackage[most]{tcolorbox}
\usepackage[dvipsnames]{xcolor}
\usepackage{enumitem}
\usepackage{fontawesome5}

\title{Not All Explanations Simulate Equally: Comparing Verbalized Feature Attributions and Self-Generated Rationales}

\author{
\textbf{Pingjun Hong \textsuperscript{\faLaptopCode\kern1pt\faGraduationCap}},
\textbf{Benjamin Roth \textsuperscript{\faLaptopCode\kern1pt\faBook}}
\\[8pt]
\textsuperscript{\faLaptopCode} Faculty of Computer Science, University of Vienna, Vienna, Austria\\
\textsuperscript{\faGraduationCap} UniVie Doctoral School Computer Science, University of Vienna, Vienna, Austria\\
\textsuperscript{\faBook} Faculty of Philological and Cultural Studies, University of Vienna, Vienna, Austria\\
\tt{
\{\href{mailto:pingjun.hong@univie.ac.at}{\textcolor{black}{pingjun.hong}},
\href{mailto:benjamin.roth@univie.ac.at}{\textcolor{black}{benjamin.roth}}\}@univie.ac.at}
}

\begin{document}
\maketitle
\begin{abstract}
Natural-language explanations are often treated as a unified interface for understanding model behavior, but different explanation sources may support simulation in different ways. This paper compares two families of explanations for question answering models: verbalized feature attributions and self-generated rationales. We evaluate them under a shared counterfactual simulation setting, using an LLM judge as predictor and measuring whether it can better predict a model's answers to follow-up questions when given its explanation. Across multiple instruction-tuned models, we analyze how explanation source, verbalization strategy, and feature granularity affect the simulatability of explanations. Our results show that explanation format and granularity affect simulatability: attribution-based explanations and self-generated rationales differ in how much they improve counterfactual prediction, with effects that vary across models and formats.
\end{abstract}

\section{Introduction}\label{sec:introduction}
Explanations for large language models (LLMs) can take many forms: input-feature-based attributions that identify important tokens or sentences \citep{Huang2023ASO,monteiro-paes-etal-2025-multi}, mechanistic accounts that describe internal circuits or representations \citep{saphra-wiegreffe-2024-mechanistic}, and natural-language rationales generated by the model itself, including post-hoc explanations and chain-of-thought (CoT) type reasoning. Despite sharing the label ``explanation'', these objects differ in their source, granularity, format, and intended use. A highlighted sentence, a token attribution map, and a fluent self-generated rationale may all appear to explain a model prediction, but they may support very different kinds of understanding.

This diversity raises a central evaluation challenge: how should we compare different types of LLM explanations? Prior work has proposed many evaluation protocols for explanations, including faithfulness to the model's decision process \citep{jacovi-goldberg-2020-towards,Turpin2023LanguageMD}, plausibility to human readers \citep{lei-etal-2016-rationalizing,Wiegreffe2021TeachMT,Agarwal2024FaithfulnessVP}, agreement with human rationales \citep{boyd-graber-etal-2022-human,brandl2026systematiccomparisonextractiveselfexplanations}, and sensitivity to input perturbations \citep{wu-etal-2021-polyjuice,ross-etal-2021-explaining,atanasova-etal-2023-faithfulness}. These protocols have led to important insights, but they often evaluate explanations along type-specific dimensions. 

We argue that \textit{simulatability} \cite{hase-bansal-2020-evaluating, hase-etal-2020-leakage} provides a useful common ground for comparing explanation types. Rather than asking only whether an explanation is plausible or faithful in isolation, simulatability asks whether the explanation helps predict the model’s behavior. This offers a functional view of explanation usefulness: an explanation is useful insofar as it improves predictive understanding of model behavior. Across explanation types, we can ask the same question: does it support accurate anticipation of the model's output under new inputs?

To answer this question for diverse explanation types, we evaluate verbalized feature attributions and self-generated rationales through counterfactual simulatability in this work. Our main findings and contributions are as follows:

\begin{figure*}[htp]
  \includegraphics[width=\textwidth]{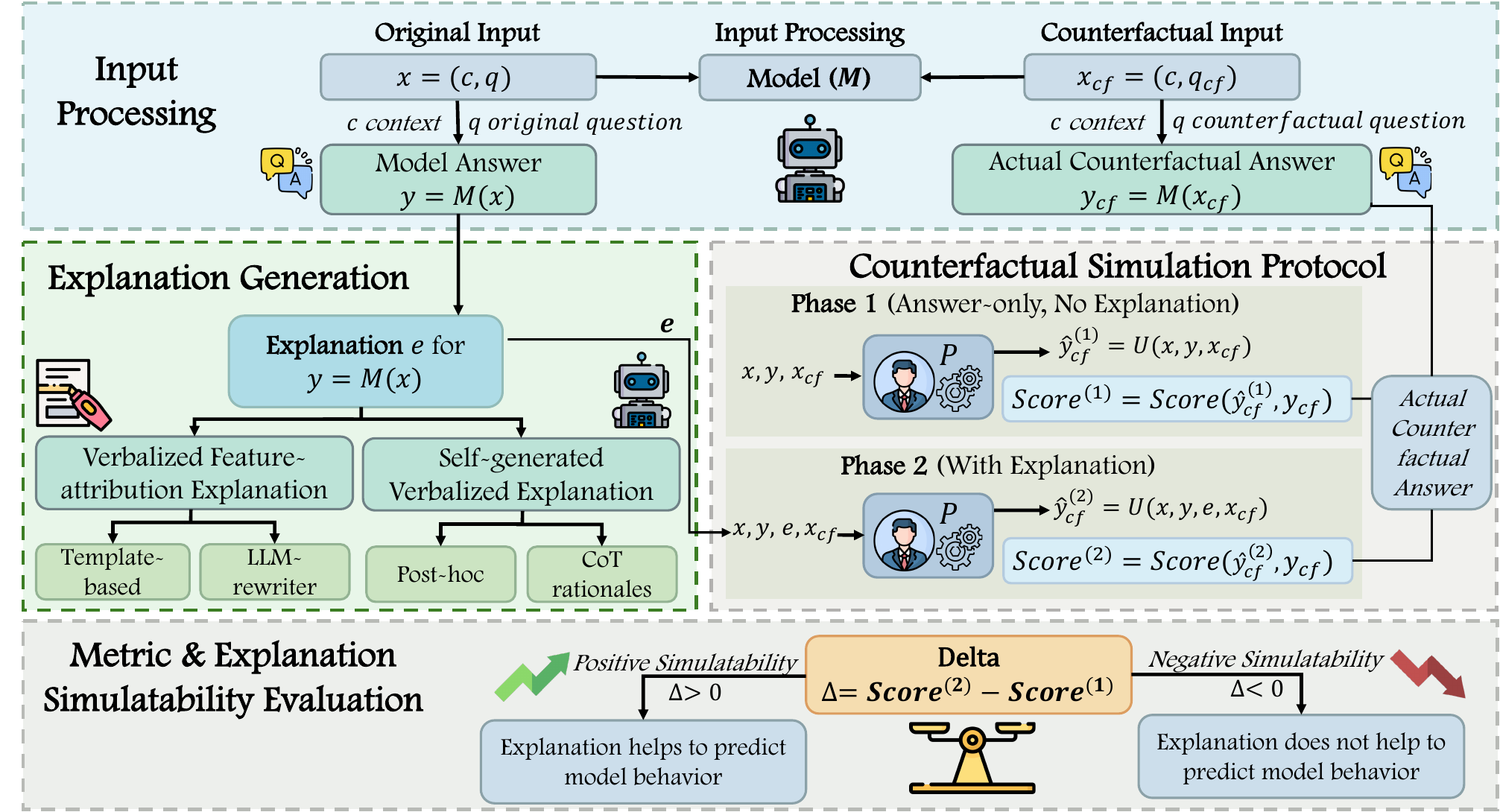}
  \caption{Overview of the proposed pipeline. Given an input $x=(c,q)$, the model produces an answer $y$ and an explanation $e$ (feature attribution or self-generated rationale). A counterfactual input $x_{cf}$ yields the true answer $y_{cf}$. In a two-phase simulation protocol, a predictor predicts $y_{cf}$ without (Phase 1) and with (Phase 2) access to $e$. \textit{Simulatability} is measured as the accuracy difference $\Delta = \text{Score}^{(2)} - \text{Score}^{(1)}$.}
  \label{fig:pipeline}
\end{figure*}

\begin{itemize}[leftmargin=*, itemsep=2pt, topsep=2pt, parsep=0pt, partopsep=0pt]
    \item We apply \textit{counterfactual simulatability} as a unified evaluation criterion for comparing different forms of LLM explanations. By measuring whether an explanation helps the prediction of model behavior on perturbed follow-up questions, this setup enables direct comparison between attribution-based explanations and self-generated rationales.
    \item We provide a controlled comparison of different verbalization methods, contrasting \textit{verbalized feature attributions} with \textit{LLM self-generated rationales} under a shared natural-language evaluation setting. For attribution-based explanations, we compare template-based and LLM-rewritten verbalizations, allowing us to study the effects of explanation source, evidence unit, and verbalization strategy.
    \item We show that \textit{explanation format and granularity have impact on simulatability}. Sentence-level attributions provide stronger simulatability gains than token-level attributions, LLM rewriting does not necessarily improve usefulness despite greater fluency, and CoT rationales provide the strongest simulation signal among the studied types. These findings highlight that the simulatability of explanations depends on both what evidence is selected and how it is presented.
\end{itemize}

\section{Counterfactual Simulatability Framework}\label{sec:simulatability-framework}

\subsection{Task Setup}
We formulate explanation evaluation as a counterfactual simulation problem. As shown in Figure~\ref{fig:pipeline}, let \(x = (c, q)\) denote the original input, where \(c\) is the context and \(q\) is the original question. A model \(M\) produces an answer \(y = M(x)\). We then define a counterfactual input \(x_{\mathrm{cf}} = (c, q_{\mathrm{cf}})\), which uses the same context but replaces \(q\) with a counterfactual question \(q_{\mathrm{cf}}\). The same model produces an actual counterfactual answer \(y_{\mathrm{cf}} = M(x_{\mathrm{cf}})\). Finally, let \(e\) denote an explanation for the model's original answer \(y\).

The goal is to evaluate whether \(e\) improves the ability to predict the model's behavior on the counterfactual input. 

\subsection{Phase-Based Simulation Protocol}

The predictor \(P\) is asked to predict the target QA model \(M\)'s answer to the counterfactual question. In Phase 1, the predictor sees the original input, the model's original answer, and the counterfactual input, but receives no explanation. In Phase 2, the predictor receives the same information plus an explanation \(e\) of the model's answer:
\begin{equation}
\begin{aligned}
\hat{y}_{\mathrm{cf}}^{(1)} &= P(x, y, x_{\mathrm{cf}}), \\
\hat{y}_{\mathrm{cf}}^{(2)} &= P(x, y, e, x_{\mathrm{cf}}).
\end{aligned}
\end{equation}

Within this protocol, explanation effect is isolated. If the explanation helps the predictor better predict \(y_{\mathrm{cf}}\), we consider it useful for simulation.

\subsection{Metrics}

We evaluate the predictions against the model's actual counterfactual answer \(y_{\mathrm{cf}}\). For each phase, we compute:
\begin{equation}
\begin{aligned}
\mathrm{Score}^{(1)} &= \mathrm{Score}(\hat{y}_{\mathrm{cf}}^{(1)}, y_{\mathrm{cf}}), \\
\mathrm{Score}^{(2)} &= \mathrm{Score}(\hat{y}_{\mathrm{cf}}^{(2)}, y_{\mathrm{cf}}), \\
\Delta &= \mathrm{Score}^{(2)} - \mathrm{Score}^{(1)}.
\end{aligned}
\end{equation}

A positive \(\Delta\) indicates improved counterfactual simulation, while a negative \(\Delta\) indicates reduced prediction accuracy.

\section{Explanation Types}
\label{sec:explanation-type}

For a systematic comparison within the simulatability framework introduced in Section~\ref{sec:simulatability-framework}, we examine two broad classes of explanations: \textit{attribution–based explanations} and \textit{self-generated natural language rationales}. Feature-attribution explanations identify which parts of the input most influenced a model output, while self-generated rationales are natural language explanations produced by the model itself. Since these two explanation types differ in format, we convert attribution scores into verbalized explanations so that they can be compared under a shared representation.

\subsection{Verbalized Feature Attributions}\label{sec:feature-attributions}
We obtain feature-based explanations using \textbf{MExGen}, a perturbation-based attribution framework for context-grounded generation \citep{monteiro-paes-etal-2025-multi}. In our setting, the resulting explanation is initially a ranked set of influential spans and we verbalize the attributed spans into explanation sentences. Motivated by prior work on verbalizing saliency maps and highlighted tokens \citep{wiegreffe-etal-2021-measuring,feldhus-etal-2023-saliency}, we test two verbalization strategies:

\paragraph{Template-based verbalization} We first verbalize the attribution results with deterministic templates. For each example, MExGen produces ranked attribution units at sentence-level and token-level. We convert these ranked units into natural language by selecting the top attributed evidence and inserting it into the following fixed templates. 

This strategy keeps verbalized explanations tightly grounded in the original attribution output by reporting only the highest-attribution input units. We evaluate three variants: \texttt{template\_sentence}, \texttt{template\_token}, which uses token-level attributions; and \texttt{template\_hybrid}, which combines the highest-ranked sentence with its top token-level units. 
These variants allow us to compare whether sentence-level, token-level, or combined evidence yields the most useful explanations. 
Implementation details are provided in Appendix~\ref{sec:mexgen-implementation}.

\begin{tcolorbox}[title=\textbf{Verbalization Template for MExGen},colback=SeaGreen!10!CornflowerBlue!10,colframe=RoyalPurple!55!Aquamarine!100!]
\textbf{\texttt{template\_sentence}:}
\small
The model mainly relies on the following sentences:
\texttt{<sentence 1>} and \texttt{<sentence 2>}.

\medskip
\textbf{\texttt{template\_token}:}

The most important tokens for the model's prediction are
\texttt{<token 1>} and \texttt{<token 2>}.

\medskip
\textbf{\texttt{template\_hybrid}:}

The model mainly relies on the following sentences: \texttt{<sentence 1>} and \texttt{<sentence 2>}.
Within this evidence, the most important tokens are \texttt{<token 1>} and \texttt{<token 2>}.

\end{tcolorbox}

\paragraph{LLM-rewriter verbalization}

The second verbalization strategy uses an LLM as a surface-level rewriter for the attribution output. Instead of generating an explanation from the original input alone, the rewriter is given the same MExGen-selected features used by the template verbalizer. The LLM is instructed to produce a single concise sentence that verbalizes these provided features, while not introducing information beyond the supplied evidence. Thus, the LLM is used only to improve fluency and readability, rather than to independently generate a new rationale. In our experiments, we use \texttt{GPT-4o-mini} as the LLM rewriter. The prompt used for LLM-rewriting is provided in Appendix~\ref{sec:llm_rewriter_prompt}.

This design separates the source of explanatory content from its linguistic form. The attribution method determines which input features are considered important, while the LLM rewriter only transforms those selected features into natural language. This allows us to evaluate whether more fluent verbalization improves simulatability without changing the underlying attribution signal.

\subsection{Self-Generated Rationales}\label{sec:self-rationales}

In addition to the attribution-based explanations, we collect self-generated explanations from the QA models themselves. These explanations are produced by prompting the same model that answers the question to also provide a natural-language rationale grounded in the provided context. We consider two variants. In the post-hoc setting, the model is first instructed to provide the answer and then to provide a concise explanation justifying its prediction from the context \citep{Camburu2018eSNLINL,Park2018MultimodalEJ}. In the CoT setting, the model is instructed to reason step by step before giving its final answer \citep{cotexplanation2022}. The generated reasoning is used as the explanation and it often exposes an explicit intermediate reasoning process leading to the final response \cite{Nye2021ShowYW,wei-jie-etal-2024-interpretable,Shen2025FaithCoTBenchBI}.

We use these self-generated explanations as a comparison for attribution-based explanations. This comparison allows us to examine whether model-produced rationales and verbalized attributions offer similar simulatability benefits, or whether they expose different signals about model behavior. The prompts used to collect self-generated rationales are provided in Appendix~\ref{sec:self_rationale_prompts}.

\section{Experimental Setup}
\label{sec:experimental-setup}

\subsection{Dataset and Counterfactual Questions}

We conduct our experiments on the \textbf{S}tanford \textbf{Q}uestion \textbf{A}nswering \textbf{D}ataset (\textbf{SQuAD}), a reading comprehension dataset consisting of crowdworker-written questions over Wikipedia passages \citep{rajpurkar-etal-2016-squad} in English. We use \textbf{SQuAD 2.0}, which includes both answerable questions, whose answers are text spans in the passage, and unanswerable questions written adversarially to resemble answerable ones \citep{rajpurkar2018knowdontknowunanswerable}. 

From SQuAD 2.0, we construct 1,204 original--counterfactual pairs by grouping examples by passage, sampling one question as the original question, and uniformly sampling another question from the same passage as the counterfactual follow-up. Appendix~\ref{sec:counterfactual-sample} shows an example pair.

\subsection{QA Models to Explain}
We evaluate explanations for three open-weight instruction-tuned models used as target QA models: \texttt{Llama-3-8B-Instruct} \citep{grattafiori2024llama3herdmodels}, \texttt{Qwen2.5-7B-Instruct} \citep{qwen2025qwen25technicalreport}, and \texttt{Mistral-7B-Instruct-v0.3} \citep{jiang2023mistral7b}.

For each model to be explained, we collect its answer to the original SQuAD 2.0 question and its answer to the counterfactual question, including \texttt{NA} when the model abstains. Explanations are generated for the model's original answer.

\subsection{Judge Model as Predictor} 
\label{sec:judge-model}

Building on prior work showing that LLMs can approximate human judgments \citep{llm-as-jusge-withmt,Bai2023BenchmarkingFM,li-etal-2025-generation}, we use an LLM judge as a predictor to evaluate \textit{simulatability}. In our experiments, the judge model is \texttt{GPT-5-mini} \citep{singh2025openaigpt5card}. The judge predicts how a target QA model would answer a counterfactual question, either without an explanation or with an explanation of the target model's original answer. Implementation details are provided in Appendix~\ref{app:llm-as-judge}.

We score the predicted answers against the target model's actual counterfactual answers using \textbf{Exact Match} and token-level \textbf{F1}. 
Exact Match measures whether the judge's prediction exactly matches the target model's answer after normalization, while F1 measures token overlap between the prediction and the target answer. 
We report $\Delta EM$ and $\Delta F1$ as the gain of the explanation condition over the no-explanation baseline.

\section{LLM Judge-based Results}\label{sec:results}


\begin{table*}[ht]
\centering
\small
\resizebox{0.8\textwidth}{!}{%
\begin{tabular}{lcccccc}
\toprule
\multirow{2}{*}{Model}
& \multicolumn{2}{c}{Template Hybrid}
& \multicolumn{2}{c}{Template Sentence}
& \multicolumn{2}{c}{Template Token} \\
\cmidrule(lr){2-3}
\cmidrule(lr){4-5}
\cmidrule(lr){6-7}
& $\Delta EM$ & $\Delta F1$
& $\Delta EM$ & $\Delta F1$
& $\Delta EM$ & $\Delta F1$ \\
\midrule
Llama-3-8B-Instruct & 8.887 & 9.181 & 16.944 & 17.474 & -0.332 & -0.411 \\
Mistral-7B-Instruct & 8.057 & 7.999 & 9.884 & 11.999 & -1.080 & -1.372 \\
Qwen2.5-7B-Instruct-v0.3 & 8.839 & 9.343 & 14.701 & 16.296 & 0.582 & 1.382  \\
\midrule
\multirow{2}{*}{Model}
& \multicolumn{2}{c}{LLM Rewriter}
& \multicolumn{2}{c}{Post-hoc Expl.}
& \multicolumn{2}{c}{CoT Rationale} \\
\cmidrule(lr){2-3}
\cmidrule(lr){4-5}
\cmidrule(lr){6-7}
& $\Delta EM$ & $\Delta F1$
& $\Delta EM$ & $\Delta F1$
& $\Delta EM$ & $\Delta F1$ \\
\midrule
Llama-3-8B-Instruct & 1.661 & 1.865 & 3.239 & 3.217 & 24.003 & 26.699 \\
Mistral-7B-Instruct  & 2.159 & 2.056 & 2.824 & 2.392 & 18.355 & 22.374\\
Qwen2.5-7B-Instruct-v0.3 & 2.409 & 2.967 & 4.485 & 3.217 & 22.259 & 24.024 \\
\bottomrule
\end{tabular}
}
\caption{
Results for the top-1 MExGen setting and single-rationale baselines. For MExGen attribution-based explanations, top-1 means that the verbalized explanation uses the highest-ranked evidence unit. For CoT and post-hoc explanations, which do not select from a ranked attribution list, we use one generated rationale per example.
}

\label{tab:main_top1}
\end{table*}

\subsection{Judge-Based Simulatability Results}
\label{sec:main_results}

We evaluate a single-explanation setting for each method. For attribution-based explanations, we verbalize the highest-ranked evidence according to the attribution score. We report $\Delta EM$ and $\Delta F1$ in Table~\ref{tab:main_top1}, so positive values indicate that the explanation improves the judge-based predictor's ability to predict model behavior under counterfactual questions. These deltas measure the \textit{simulatability} of the explanation. A random feature baseline is provided in Appendix~\ref{sec:comparision-random-selection}. Our key observations are:

\paragraph{Sentence-level attributions are the most useful attribution-based explanations.}
Among verbalized feature-attribution explanations, sentence-level evidence provides the strongest gains. \texttt{template\_hybrid} also improves performance across all models, but by a smaller margin. This suggests that, for in-context QA, providing the most influential sentence gives the LLM judge a clearer signal about the model's behavior than combining sentence-level with token-level features.

\paragraph{Word-level spans provide insufficient context.}
In contrast, \texttt{template\_token} produces near-zero or negative changes: individual word-level spans, despite receiving high attribution scores, often lack the broader context needed to support simulation.

\paragraph{Fluency does not guarantee simulatability.}
The results of implementing the LLM Rewriter show that fluency does not necessarily translate into simulatability. Despite rewriting the top-1 attribution evidence into more fluent free-text, LLM Rewriter yields only very small gains across all models. This suggests that \textit{simulatability} depends more on the provided information than on surface fluency. In our setting, template-based sentence explanations are substantially more helpful for simulation.

\paragraph{CoT rationales provide the strongest simulation signal.}
Self-generated rationales show a contrast between post-hoc explanations and CoT rationales. Post-hoc explanations provide modest improvements across models, while CoT rationales produce the largest gains in the table. This indicates that not all LLM self-generated explanations are equally useful ,and step-by-step rationales provide much stronger simulation signals. 

\paragraph{Additional Judge Robustness.}
As a robustness check, we also evaluate the \texttt{Llama-3} explanations using \texttt{DeepSeek-R1} \citep{Guo_2025} as an additional judge. While the two judges do not always agree exactly on free-form predictions, the high-level pattern remains similar: CoT produces the largest gains, with $17.940$ in $\Delta EM$ and $22.519$ in $\Delta F1$, and sentence-level evidence is the strongest feature-attribution verbalization. LLM-rewritten explanations improve even less under \texttt{DeepSeek-R1}, with a $\Delta EM$ of $0.581$ and a $\Delta F1$ of $0.921$, whereas post-hoc explanations remain weak. The full results and inter-judge agreement analysis are in Appendix~\ref{app:r1-judge}. 

\subsection{Judge-Based Sample-level Analysis}
\label{sec:sample_level_shifts}

\begin{figure}[ht]
  \includegraphics[width=\columnwidth]{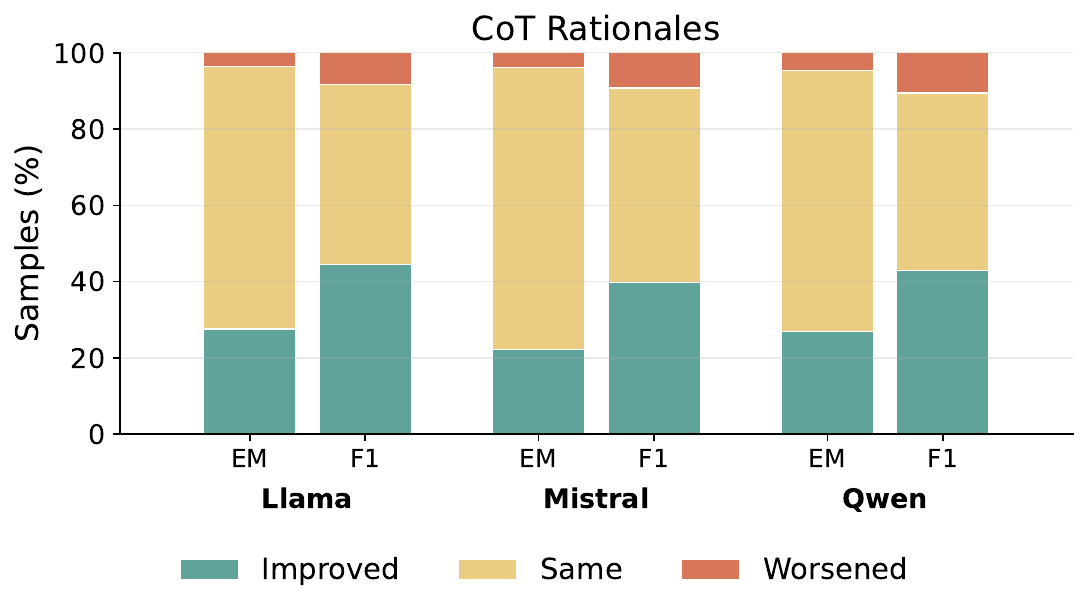}
  \caption{
    Sample-level shifts in counterfactual simulation accuracy with CoT explanations. Each subplot corresponds to one explanation condition, and each target model is represented by two stacked bars.
    }
  \label{fig:sample-level-shifts}
\end{figure}

While Table~\ref{tab:main_top1} reports aggregate changes in EM and F1, Figure~\ref{fig:sample-level-shifts} decomposes those changes into three categories: examples where adding the explanation improves the accuracy of the prediction, examples where the prediction remains unchanged, and examples where the explanation worsens the prediction relative to the no-explanation baseline. The full results and raw counts for all explanation types are provided in Appendix~\ref{app:sample_level_shift_counts}.

The sample-level results show that the strongest methods improve aggregate performance by shifting a substantial number of examples from incorrect to correct predictions. To better understand these shifts, we inspect individual CoT cases. Table~\ref{tab:qualitative-cot} shows one example where the rationale improves simulation and one where it hurts simulation. In the improved example, the no-explanation judge predicts a concrete founding date for the \textit{Office of Eastern Medicine}. The CoT rationale, however, highlights a mismatch: the context mentions the \textit{Western Medicine}, but not the \textit{Eastern Medicine}. With this additional signal, the judge changes its prediction to \texttt{NA}, matching the target model's answer. The worsened case illustrates the opposite: the CoT rationale causes the judge to over-correct to NA, despite the correct answer being \textit{El Temür}.

\begin{table}[t]
\centering
\resizebox{\linewidth}{!}{%
\begin{tabular}{p{0.18\linewidth} p{0.82\linewidth}}
\toprule
Case & Example \\
\midrule
\textbf{\textcolor{PineGreen}{Improved}} &
\textbf{Counterfactual question:} When was the Office of Eastern Medicine founded? \newline
\textbf{CoT:} The context does not mention the \textit{Office of Eastern Medicine}. Instead, it mentions the \textit{Office of Western Medicine}, which was founded by Jesus the Interpreter in 1263. \newline
\textbf{No expl.:} It was founded in 1890; \textbf{With CoT:} \texttt{NA}; \textbf{Target output:} \texttt{NA} \\
\midrule
\textbf{\textcolor{Bittersweet}{Worsened}} &
\textbf{Counterfactual question:} Who was thought to have killed Tugh Temur? \newline
\textbf{CoT:} Kusala died suddenly after a banquet with Tugh Temür, and El Temür \textit{supposedly} killed him with poison. \newline
\textbf{No expl.:} El Temür; \textbf{With CoT:} \texttt{NA}; \textbf{Target output:} El Temür \\
\bottomrule
\end{tabular}%
}
\caption{Qualitative examples of sample-level shifts for CoT rationales.}
\label{tab:qualitative-cot}
\end{table}

\section{Deeper Analysis of Explanation Simulatability}
\label{sec:deeper-analysis}

The results in Section~\ref{sec:results} raise further questions about what drives simulatability and whether the findings generalize. We examine three of them: attribution evidence quantity (\S\ref{sec:topk_ablation}), the source of CoT's gains (\S\ref{sec:cross-model}), and consistency across evaluation paradigms (\S\ref{sec:teacher-student}).

\subsection{Effect of the Number of Attributed Features}
\label{sec:topk_ablation}

\begin{figure}[h]
  \includegraphics[width=\columnwidth]{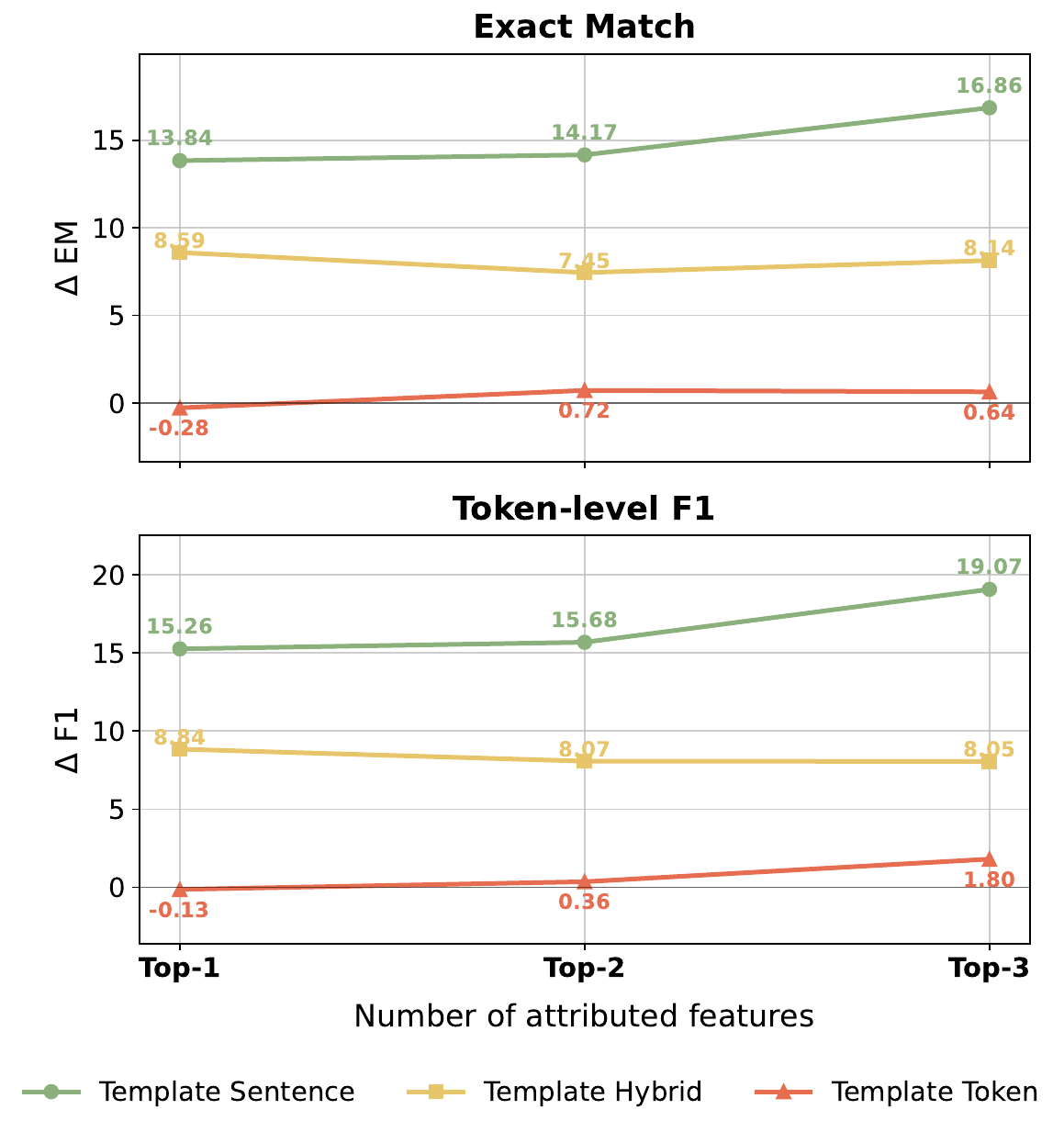}
  \caption{
    Top-$k$ ablation for template-based verbalized attribution explanations. We vary the number of MExGen-selected features included in the explanation from top-1 to top-3.
    }
  \label{fig:topk-ablation}
\end{figure}

Figure~\ref{fig:topk-ablation} shows how the measured simulatability changes as we include more MExGen-selected features. We report average $\Delta EM$ and $\Delta F1$ across target models as the number of MExGen-selected features increases from top-1 to top-3. The full per-model results are shown in Appendix~\ref{app:topk_ablation_per_model}.

Overall, \texttt{template\_sentence} explanations remain the strongest attribution-based verbalization across all values of $k$. Increasing from top-1 to top-3 improves the average gain from $+13.843$ to $+16.860$ in EM and from $+15.256$ to $+19.066$ in F1. This suggests that additional sentence-level evidence can provide useful context for the prediction of model behavior, rather than simply adding noise.

Explanations based on \texttt{template\_hybrid} show a different pattern. Their average gains remain positive, but they do not improve as more features are added. The EM gain decreases from $+8.594$ at top-1 to $+7.447$ at top-2, then recovers to $+8.139$ at top-3; F1 follows a similar trend.

Token-level explanations remain the weakest. Their average $\Delta EM$ and $\Delta F1$ are close to zero for top-1 and top-2, with only a small improvement at top-3. Even when more tokens are included, token-level verbalizations provide much less simulation benefit than sentence-level explanations. This reinforces our main finding that the unit of evidence matters: for in-context QA, sentence-level attributions offer a more useful basis for predicting model behavior than isolated token-level features.

\subsection{Cross-Model CoT Transfer}
\label{sec:cross-model}

To examine whether CoT's gains reflect model-specific signals or task-relevant information that generalize across models, we provide one model's CoT to the judge when predicting a \textit{different} model's answers.

\begin{figure}[h]
  \includegraphics[width=\columnwidth]{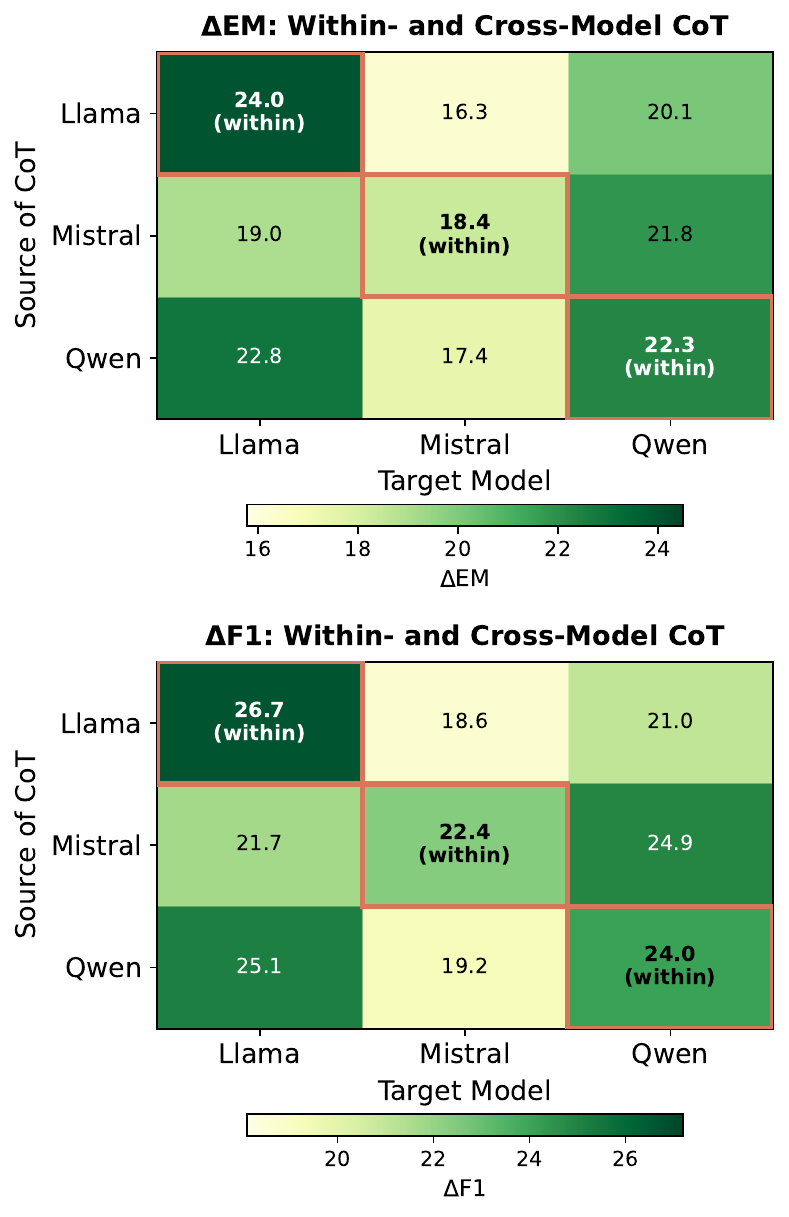}
  \caption{
    $\Delta EM$ and $\Delta F1$ for within- and cross-model CoT transfer. Rows indicate the source model whose CoT rationale is provided to the judge; columns indicate the target QA model. Diagonal entries (\textcolor{Bittersweet}{red border}) correspond to the standard within-model setting.
    }
  \label{fig:cross_model}
\end{figure}

As shown in Figure~\ref{fig:cross_model}, cross-model transfer yields consistently positive $\Delta$, indicating that CoT rationales carry substantial task-relevant information that generalizes across models sharing the same context. At the same time when comparing each column, cross-model gains remain consistently lower than the within-model condition: for example, Llama's CoT achieves $\Delta EM$ of 24.0 within-model but averages only 18.2 when used to predict other models. This gap, while modest, suggests that CoT also encodes model-specific signals that do not fully transfer. Task-relevant reasoning thus appears to be the dominant component of CoT's simulatability gains, with model-specific information playing a secondary but non-negligible role. This interpretation is further corroborated by the teacher--student results in Section~\ref{sec:teacher-student}, where CoT remains the only explanation type to consistently yield positive training gains, which is harder to attribute to answer leakage alone.

\subsection{Teacher-Student Simulatability Evaluation}
\label{sec:teacher-student}

As an alternative simulatability evaluation, we train a smaller student model to approximate the answer behavior of the target QA model (the teacher), following \citet{pruthi_evaluating_2022}. We compare two variants: an answer-supervised student optimized only on teacher-answer tokens, and an explanation-supervised student additionally optimized on teacher explanation. Unlike the judge-based setting, this setup tests whether explanation supervision can be internalized through training rather than merely exploited in context. Figure~\ref{fig:teacher-student} illustrates the setup; details are in Appendix~\ref{sec:teacher-student-setup}.
\begin{figure}[h]
    \centering
    \includegraphics[width=\columnwidth]{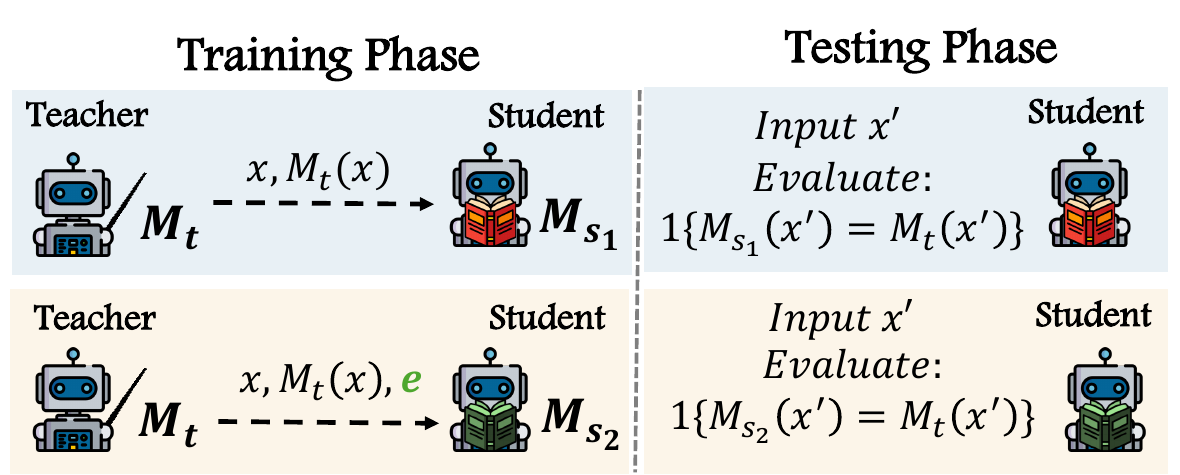}
    \caption{Teacher-student simulatability evaluation.}

    \label{fig:teacher-student}
\end{figure}

After training, we evaluate both students using the counterfactual context and question as input. The Student's predictions are compared against the teacher's actual counterfactual answers. We report the improvement of the explanation-input student over the base student as $\Delta EM$ and $\Delta F1$. 

To examine whether teacher-student evaluation produces consistent conclusions with judge-based evaluation, we compute the Spearman rank correlation between their improvement scores. 
For each explanation type we compute
\begin{equation}
\rho = \mathrm{Spearman}\left(\{\Delta^{\mathrm{Judge}}_i\}_{i=1}^{N}, \{\Delta^{\mathrm{ST}}_i\}_{i=1}^{N}\right),
\end{equation}
where each $i$ corresponds to an aligned model--explanation--metric entry. 

\begin{figure}[ht]
    \centering
    \includegraphics[width=\columnwidth]{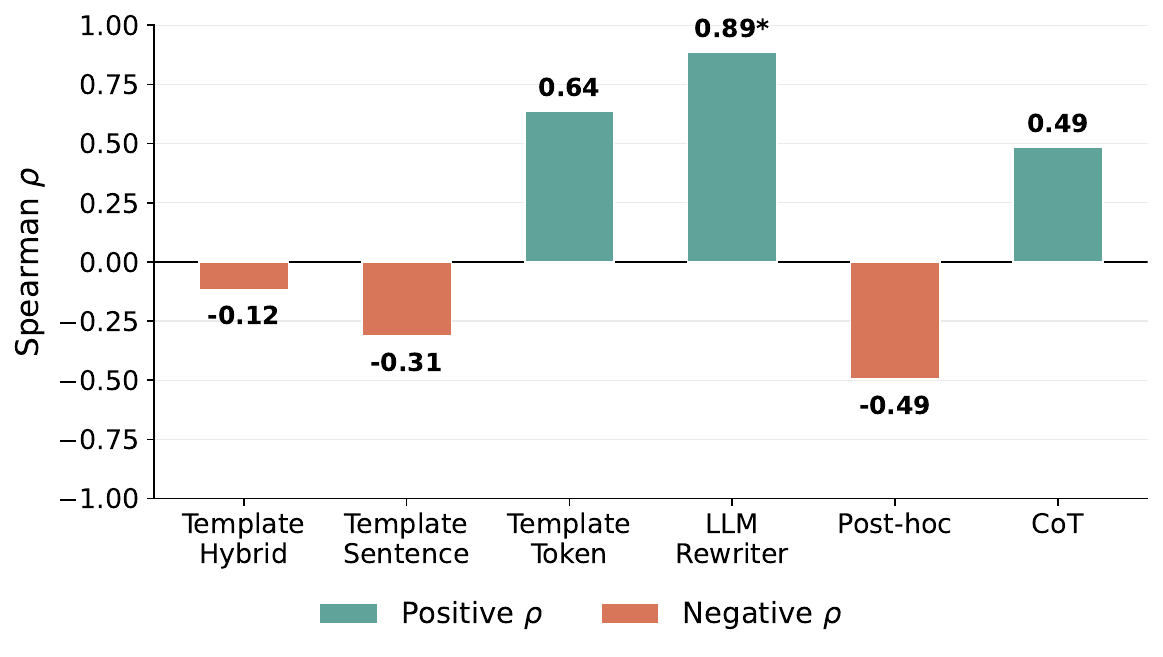}
    \caption{Spearman rank correlation ($\rho$) between judge-based and teacher--student evaluation paradigms, computed per explanation type over models and metrics.}
    \label{fig:spearman}
\end{figure}

As shown in Figure~\ref{fig:spearman}, rank agreement varies substantially across explanation types. CoT rationales show the strongest positive $\rho$, indicating that they are consistent for both protocols. In contrast, template-based attribution explanations show weaker or negative correlations, suggesting that gains observed under judge-based evaluation do not transfer to the teacher--student setting. This discrepancy reflects a fundamental difference between the two evaluation paradigms: judge-based evaluation measures whether an explanation improves immediate in-context prediction, whereas teacher--student evaluation tests whether the explanation provides a learnable supervision signal.

\section{Related Work}\label{sec:related-work}

\subsection{Feature Attribution and Verbalized Explanations}

Feature attribution methods explain model behavior by identifying which parts of the input influence a prediction. For classification models, perturbation-based attribution methods such as LIME \citep{ribeiro-etal-2016-trust} and SHAP \citep{Lundberg2017AUA} have been widely used to assign importance scores to input spans. Extending these methods to generative models is less straightforward, because the output is a sequence rather than a label \citep{mosca-etal-2022-shap,enouen-etal-2024-textgenshap}. MExGen approaches this problem by identifying influential input features by perturbing input units and observing changes in the output \citep{monteiro-paes-etal-2025-multi}. This makes it suitable for explaining in-context QA task, where the goal is to identify which context units support the answer. However, raw attribution outputs are not in natural-language explanations, and the selected spans must still be verbalized before they can be used in real-life scenarios.

Prior works also study another type of model explanations: self-explanations, where models are prompted to explain their own predictions \citep{Camburu2018eSNLINL, huang2023largelanguagemodelsexplain, kroeger2024incontextexplainersharnessingllms, madsen-etal-2024-self}. Although self-explanations are fluent and easy to elicit, several studies have questioned whether they faithfully reflect the model's actual decision process, especially in in-context classification and QA settings \citep{huang2023largelanguagemodelsexplain, madsen-etal-2024-self, Fragkathoulas_2024}. This motivates comparing externally derived attribution explanations with rationales produced by the model itself.

In our work, we use MExGen to obtain feature-attribution explanations for QA models and then convert the selected features into natural language. 

\subsection{Simulation-Based Evaluation of Explanations}

Explanations can be evaluated by whether they help users anticipate how a model will behave on new or modified inputs. \textit{Simulatability} formalizes this criterion by measuring how well a model's outputs can be predicted when its explanations are available \citep{doshivelez2017towards}. Prior work distinguishes between \textit{forward simulation}, which involves predicting a model's output from an input and explanation, and \textit{counterfactual simulation}, which involves predicting the model's behavior under a perturbed input after observing the original input, output, and explanation. \citep{hase-bansal-2020-evaluating,poche-etal-2025-consim}.

Simulation-based evaluations have been used to assess whether explanations improve the ability to anticipate model outputs \citep{ribeiro-etal-2016-trust,chandrasekaran-etal-2018-explanations,nguyen-2018-comparing,chen2023modelsexplainthemselvescounterfactual,limpijankit-etal-2025-counterfactual,Hong2026DoLS,mayne2026positivecasefaithfulnessllm}. More broadly, these evaluations relate to the goal of understanding model behavior, including when the model is likely to be correct and where its limitations or biases may arise \citep{miller_explanation_2018,bansal_beyond_2019}.

We adopt counterfactual simulation as the evaluation setting for LLM-explanations, examining whether verbalized feature attributions and self-generated rationales help predict model behavior.

\section{Conclusion}
\label{sec:conclusion}

This work compares natural-language explanations from different sources through the lens of \textit{simulatability}, evaluating whether they help an LLM judge predict how a target QA model answers counterfactual questions. Our results show that explanation form matters: sentence-level verbalizations are the most useful attribution-based explanations, while word-level spans provide weak or noisy signals. Template-based verbalizations preserve attribution evidence more effectively than LLM-rewritten alternatives despite lower fluency, and CoT rationales provide substantially larger gains than post-hoc explanations. Further analysis shows that sentence-level attribution gains increase with the number of selected features, CoT carries substantial task-relevant information that transfers across models while retaining model-specific signals, and judge-based and teacher--student evaluations agree most strongly for CoT but diverge for attribution-based explanations, suggesting the two paradigms capture different aspects of explanation utility.

Overall, our findings suggest that explanation evaluation should account for both the source of the explanation and how it is verbalized. When explanations are intended to support anticipation of model behavior, \textit{simulatability} offers a practical behavioral criterion. Our comparison provides guidance for choosing explanation formats in QA settings and highlights the need to evaluate the utility of self-generated rationales against attribution-grounded alternatives, rather than assuming that all natural-language explanations serve the same role.

\section*{Limitations}

Our primary evaluation relies on an LLM judge (GPT-5-mini) to measure counterfactual simulatability. We chose an LLM-as-a-judge approach because alternative automated metrics, such as perturbation sensitivity or learning-based simulation, either fail to capture functional comprehension or are prone to overfitting on spurious features. To mitigate the inherent biases of any single LLM, we validated our judge-based findings against a teacher-student framework to provide a more rigorous, multi-paradigm view of explanation usefulness. Robustness checks using DeepSeek-R1 confirm that our macro-level conclusions, most notably the superior simulatability of Chain-of-Thought rationales, hold consistent across different evaluator architectures. It is an open question, however, under which conditions and how closely LLM-predicted simulatability aligns with actual human cognitive processing.

Our experiments are currently constrained to the SQuAD 2.0 dataset. We selected this dataset because its unique structure, which features multiple questions per paragraph and adversarially written unanswerable queries, provides an ideal, highly controlled environment for testing counterfactual simulatability and the transfer of explanation information within shared contexts. It remains an open question how well these simulatability patterns generalize to open-ended generative tasks, complex reasoning, and abstractive summarization, where models rely primarily on internal parametric knowledge rather than external text.

Results for feature-attribution usefulness are tied to our reliance on the MExGen implementation of SHAP and fixed verbalization templates. We utilized MExGen because traditional attribution implementations are predominantly designed for classifiers; MExGen remains the most established framework for generative language models and offers theoretical optimality as an instantiation of SHAP. We omitted earlier gradient- or attention-based methods due to a lack of standardized generative implementations and the ongoing debate regarding whether attention constitutes a faithful explanation. To isolate this attribution signal from surface-level fluency, we employed rigid templates, a choice validated by our LLM-rewriter ablation, which confirmed that evidence granularity drives simulatability more than linguistic formulation.

Finally, our comparison deliberately isolates counterfactual simulatability as a functional lens for evaluating explanations, rather than offering a comprehensive measure of explanation quality. An explanation that improves the simulated prediction of model behavior may still differ along other critical dimensions, such as algorithmic faithfulness, user helpfulness, or trust calibration.

\section*{Acknowledgments} 

\paragraph{Ethical considerations.} We do not foresee any ethical concerns associated with this work. All analyses were conducted using publicly available datasets and models. No private or sensitive information was used. 

\paragraph{Reproducibility and responsible release.}
We will release all code, prompts, evaluation scripts, and model-training configurations at publication time to support reproducibility. 

\paragraph{Use of AI Assistants.} The authors acknowledge the use of ChatGPT for correcting grammatical errors, enhancing the coherence of the manuscripts, and providing coding assistance.

\bibliography{custom}
\bibliographystyle{acl_natbib}

\clearpage

\appendix
\input{appendix}

\end{document}

%% file: appendix.tex
\section{MExGen Implementation}
\label{sec:mexgen-implementation}

We implemented MExGen explanations using the ICX360 library for context-grounded question answering \citep{monteiro-paes-etal-2025-multi,wei2025icx360incontextexplainability360}. For each SQuAD example, the input is formatted as:

\begin{quote}
\texttt{Context: <context>, Question: <question>, Answer:}
\end{quote}

The model first generated a short answer using only the given context. We then apply MExGen with the probability scalarizer to attribute the generated answer back to the input context. Our implementation first compute sentence-level attributions, selecte the top-ranked sentence, and then perform a second mixed-level explanation over that selected sentence to obtain finer-grained word-level evidence.

In the main experiments, we run the three models under explanation: \texttt{Llama-3-8B-Instruct}, \texttt{Qwen2.5-7B-Instruct}, and \texttt{Mistral-7B-Instruct-v0.3}. For all models, we used L-SHAP as the MExGen explainer, CUDA execution with bfloat16 precision, and a maximum generation length of 128 tokens.

For each example, the pipeline saves the model answer, sentence-level attribution scores, and mixed-level attribution scores.

\section{Prompt for LLM Rewriter}
\label{sec:llm_rewriter_prompt}

We use an LLM rewriter to convert selected MExGen attribution features into fluent natural-language explanations. The rewriter receives the model answer, the original question, and the selected top-$k$ attribution features. The prompt below shows the instruction template used.

\begin{tcolorbox}[
    colback=gray!5,
    colframe=gray!60,
    breakable,
    title=LLM Rewriter Prompt,
    boxsep=2pt,
    left=4pt,
    right=4pt,
    top=4pt,
    bottom=4pt
]
\small
\setlength{\parskip}{2pt}
You rewrite feature-based explanations for extractive question answering.

\textbf{Rules:} Output exactly one concise English sentence. Do not add facts that are not present in the evidence. Do not mention scores, top-k, JSON, prompts, or templates. Use wording such as ``The model appears to rely on...'' when helpful. Start the final line with ``EXPLANATION:''.

\textbf{Question:} \{question\}

\textbf{Model answer:} \{answer\}

\textbf{Selected top-k level:} top\{k\}

\textbf{Important sentences:}

1. \{selected sentence 1\}

2. \{selected sentence 2, if used\}

3. \{selected sentence 3, if used\}

\textbf{Important tokens/phrases:}

1. \{selected token or phrase 1\}

2. \{selected token or phrase 2, if used\}

3. \{selected token or phrase 3, if used\}

\textbf{Task:} 

Write one fluent explanation of why the model predicted the answer, grounding it in the important sentence evidence and highlighting the important tokens or phrases.

\textbf{Output:}
\begin{verbatim}
EXPLANATION: 
<one concise rewritten explanation>
\end{verbatim}
\end{tcolorbox}

\section{Prompts for Self-Generated Explanations}
\label{sec:self_rationale_prompts}

We use two prompting strategies to collect self-generated rationales from the QA models: post-hoc explanations and CoT rationales. The prompts below show the system-level instructions used for each setting.

\subsection{Post-hoc Rationale Prompt}

\begin{tcolorbox}[
    colback=gray!5,
    colframe=gray!60,
    breakable,
    title=Post-hoc Prompt,
    boxsep=2pt,
    left=4pt,
    right=4pt,
    top=4pt,
    bottom=4pt
]
\small
\setlength{\parskip}{2pt}
You are a QA assistant. Use ONLY the provided context.

\textbf{Task:} Predict the answer. Provide a post-hoc self-explanation that grounds the reasoning in the context, explaining why the answer follows from the context.

\textbf{Output format (STRICT JSON, no extra text):}
\begin{verbatim}
{"answer": 
"<short exact span copied from context>",
 "explanation": 
 "<1-3 sentences explaining the reasoning>"}
\end{verbatim}

\textbf{Rules:} The explanation must be based only on information in the context. Answer NA if the question is unanswerable based on the information in the context. Keep the explanation concise and focused on the causal/logical link.
\end{tcolorbox}

\subsection{Chain-of-Thought Rationale Prompt}

\begin{tcolorbox}[
    colback=gray!5,
    colframe=gray!60,
    breakable,
    title=Chain-of-Thought Prompt,
    boxsep=2pt,
    left=4pt,
    right=4pt,
    top=4pt,
    bottom=4pt
]
\small
\setlength{\parskip}{2pt}
You are a QA assistant. Use ONLY the provided context.

\textbf{Task:} Think step by step BEFORE giving your final answer.

\textbf{Output format (STRICT JSON, no extra text):}
\begin{verbatim}
{"reasoning": 
"<step-by-step reasoning from context>",
 "answer": 
 "<short exact span copied from context>"}
\end{verbatim}

\textbf{Rules:} Reasoning must cite evidence from the context step by step. Answer NA if the question is unanswerable based on the context. Keep each reasoning step concise and logically connected.
\end{tcolorbox}

\section{Explanation Statistics}
\label{app:explanation-statistics}

Table~\ref{tab:explanation-length} reports the average explanation length across explanation types and models, followed by two items paired with six explanation variants.

\begin{table}[t]
\centering
\small
\resizebox{\columnwidth}{!}{
\begin{tabular}{llrr}
\toprule
Explanation & Model & Avg. Words & Avg. Chars \\
\midrule
Template Hybrid & Llama-3-8B-Instruct  & 45.39 & 292.9 \\
Template Token & Llama-3-8B-Instruct  & 10.00 & 64.5 \\
Template Sentence & Llama-3-8B-Instruct  & 36.39 & 230.4 \\
LLM Rewriter & Llama-3-8B-Instruct  & 30.52 & 194.7 \\
Post-hoc Expl.& Llama-3-8B-Instruct  & 26.01 & 162.9 \\
CoT Rationale & Llama-3-8B-Instruct  & 31.26 & 196.4 \\
\midrule
Template Hybrid & Mistral-7B-Instruct & 46.47 & 300.0 \\
Template Token & Mistral-7B-Instruct & 10.00 & 64.7 \\
Template Sentence & Mistral-7B-Instruct & 37.47 & 237.4 \\
LLM Rewriter & Mistral-7B-Instruct & 31.13 & 198.2 \\
Post-hoc Expl. & Mistral-7B-Instruct & 22.12 & 137.4 \\
CoT Rationale & Mistral-7B-Instruct & 32.85 & 210.6 \\
\midrule
Template Hybrid & Qwen2.5-7B-Instruct-v0.3 & 44.71 & 288.7 \\
Template Token & Qwen2.5-7B-Instruct-v0.3 & 10.00 & 64.5 \\
Template Sentence & Qwen2.5-7B-Instruct-v0.3 & 35.71 & 226.2 \\
LLM Rewriter & Qwen2.5-7B-Instruct-v0.3 & 30.54 & 194.5 \\
Post-hoc Expl. & Qwen2.5-7B-Instruct-v0.3 & 20.55 & 130.7 \\
CoT Rationale & Qwen2.5-7B-Instruct-v0.3 & 35.69 & 230.1 \\
\bottomrule
\end{tabular}
}
\caption{Average explanation length by explanation type and model.}
\label{tab:explanation-length}
\end{table}

\begin{tcolorbox}[title=\textbf{Example 1},colback=SeaGreen!10!CornflowerBlue!10,colframe=RoyalPurple!55!Aquamarine!100!]
\textbf{Question:} Who patented a high-pressure compound engine in 1805?

\textbf{Answer:} Arthur Woolf.

\textbf{Template Hybrid:} The model mainly relies on the sentence ``A method to lessen the magnitude of this heating and cooling was invented in 1804 by British engineer Arthur Woolf, who patented his Woolf high-pressure compound engine in 1805.''. Within this evidence, the most important tokens are ``Arthur''.

\textbf{Template Token:} The most important token for the model's prediction is ``Arthur''.

\textbf{Template Sentence:} The model mainly relies on the sentence ``A method to lessen the magnitude of this heating and cooling was invented in 1804 by British engineer Arthur Woolf, who patented his Woolf high-pressure compound engine in 1805.''.

\textbf{LLM Rewriter:} The model appears to rely on the information that British engineer Arthur Woolf patented his high-pressure compound engine in 1805, as stated in the important sentence.

\textbf{Post-hoc Expl:} The text states that Arthur Woolf patented his Woolf high-pressure compound engine in 1805, directly linking him to the patent.

\textbf{CoT Rationale:} The context states that Arthur Woolf patented his Woolf high-pressure compound engine in 1805, not 1804.
\end{tcolorbox}

\begin{tcolorbox}[title=\textbf{Example 2},colback=SeaGreen!10!CornflowerBlue!10,colframe=RoyalPurple!55!Aquamarine!100!]
\textbf{Question:} What kinds of literature were Social Darwinism and theories of race?

\textbf{Answer:} pseudo-sciences

\textbf{Template Hybrid:} The model mainly relies on the sentence ``The pseudo-sciences of Social Darwinism and theories of race formed an ideological underpinning during this time.''. Within this evidence, the most important tokens are ``pseudo''.

\textbf{Template Token:} The most important token for the model's prediction is ``pseudo''.

\textbf{Template Sentence:} The model mainly relies on the sentence ``The pseudo-sciences of Social Darwinism and theories of race formed an ideological underpinning during this time.''.

\textbf{LLM Rewriter:} The model appears to rely on the characterization of Social Darwinism and theories of race as pseudo-sciences, which served as an ideological foundation during that period.

\textbf{Post-hoc Expl.:} Social Darwinism and theories of race are referred to as ``pseudo-sciences'' because they are mentioned in the context as ``pseudo-sciences''.

\textbf{CoT Rationale:} Social Darwinism and theories of race are described as ``pseudo-sciences'' in the context, indicating that they are not recognized scientific theories or disciplines.
\end{tcolorbox}

Overall, the explanation types differ substantially in both length and specificity. 
Template token explanations are the shortest because they only verbalize the most important token or phrase, while sentence and hybrid templates are longer because they include sentence-level evidence. LLM-rewritten explanations are more compact than the corresponding templates. 
Post-hoc and CoT explanations vary more in coverage, with some items receiving empty or incomplete explanations, but when available they often provide more interpretive reasoning than the template-based explanations. 

\section{Example of Original--Counterfactual Pair from SQuAD 2.0}
\label{sec:counterfactual-sample}

The appendix shows one example from the SQuAD 2.0 counterfactual set. The original and counterfactual questions share the same passage, but ask for different pieces of information. In this example, the original question is an adversarial unanswerable variant because the passage states that the Islamic State proclaimed itself a caliphate in 2014, not 2015, whereas the counterfactual question asks for an answerable description from the same passage.

\begin{tcolorbox}[title=\textbf{Original–counterfactual question pair},colback=SeaGreen!10!CornflowerBlue!10,colframe=RoyalPurple!55!Aquamarine!100!]
\small
\textbf{Title:} Islamism

\medskip
\textbf{Passage:} ``The Islamic State'', formerly known as the ``Islamic State of Iraq and the Levant'' and before that as the ``Islamic State of Iraq'', (and called the acronym Daesh by its many detractors), is a Wahhabi/Salafi jihadist extremist militant group which is led by and mainly composed of Sunni Arabs from Iraq and Syria. In 2014, the group proclaimed itself a caliphate, with religious, political and military authority over all Muslims worldwide. As of March 2015[update], it had control over territory occupied by ten million people in Iraq and Syria, and has nominal control over small areas of Libya, Nigeria and Afghanistan. (While a self-described state, it lacks international recognition.) The group also operates or has affiliates in other parts of the world, including North Africa and South Asia.

\medskip
\textbf{Original question:} What did the Islamic State proclaim itself in 2015?

\medskip
\textbf{Counterfactual question:} What type of group is The Islamic State?

\medskip
\textbf{Number of counterfactual candidates:} 9
\end{tcolorbox}

\section{LLM-as-a-Judge Evaluation Setup}
\label{app:llm-as-judge}

\begin{table*}[ht]
\centering
\small
\resizebox{\textwidth}{!}{
\begin{tabular}{lcccccc}
\toprule
Model
& Template Hybrid
& Template Sentence
& Template Token
& LLM Rewriter
& Post-hoc Expl.
& CoT Rationale \\
\midrule
\multicolumn{7}{c}{\textbf{Exact Match: Improved / Same / Worsened}} \\
\midrule
Llama-3-8B-Instruct
& 167 / 981 / 58
& 265 / 878 / 61
& 58 / 1084 / 62
& 57 / 1110 / 37
& 90 / 1063 / 51
& 332 / 829 / 43 \\
Mistral-7B-Instruct 
& 164 / 973 / 67
& 215 / 893 / 96
& 53 / 1085 / 66
& 52 / 1126 / 26
& 99 / 1040 / 65
& 267 / 891 / 46 \\
Qwen2.5-7B-Instruct-v0.3
& 169 / 967 / 68
& 271 / 839 / 94
& 64 / 1083 / 57
& 62 / 1109 / 33
& 112 / 1034 / 58
& 324 / 824 / 56 \\
\midrule
\multicolumn{7}{c}{\textbf{F1: Improved / Same / Worsened}} \\
\midrule
Llama-3-8B-Instruct
& 267 / 814 / 123
& 402 / 694 / 108
& 153 / 889 / 162
& 221 / 755 / 228
& 212 / 839 / 153
& 535 / 570 / 99 \\
Mistral-7B-Instruct
& 265 / 796 / 143
& 359 / 688 / 157
& 125 / 935 / 144
& 224 / 766 / 214
& 170 / 889 / 145
& 482 / 620 / 112 \\
Qwen2.5-7B-Instruct-v0.3
& 295 / 762 / 147
& 442 / 615 / 147
& 181 / 887 / 136
& 237 / 753 / 214
& 212 / 836 / 156
& 517 / 560 / 127 \\
\bottomrule
\end{tabular}
}
\caption{
Raw sample-level shift counts underlying Figure~\ref{fig:sample-level-shifts}. Each entry is reported as Improved / Same / Worsened relative to the no-explanation baseline.
}
\label{tab:sample_shift_counts}
\end{table*}

We use an LLM-as-a-judge protocol to evaluate whether a natural-language explanation helps predict a model's behavior under counterfactual questions. The evaluation is framed as a counterfactual simulation task. For each example, the judge is given a starter question, the model's answer to that starter question, and a follow-up question written under the same context. The judge is asked to predict the answer that the same model would give to the follow-up question. 

We evaluate two phases. In Phase 1, the judge observes only the starter question, the model's starter answer, and the counterfactual follow-up question. In Phase 2, the judge receives the same inputs plus the explanation associated with the starter answer. 

\paragraph{Judge Prompt.}
The system message is:

\begin{tcolorbox}[
    colback=gray!5,
    colframe=gray!60,
    breakable,
    title=LLM-as-a-Judge System Prompt,
    boxsep=2pt,
    left=4pt,
    right=4pt,
    top=4pt,
    bottom=4pt
]
\small
\setlength{\parskip}{2pt}
You are a careful judge that follows instructions exactly.
\end{tcolorbox}

The user prompt begins with the following instruction:

\begin{tcolorbox}[
    colback=gray!5,
    colframe=gray!60,
    breakable,
    title=LLM-as-a-Judge Instruction Prompt,
    boxsep=2pt,
    left=4pt,
    right=4pt,
    top=4pt,
    bottom=4pt
]
\small
\setlength{\parskip}{2pt}
Human: In the task below, you will read a starter question and a model's answer to that question. Then you will read a follow-up question under the same context. Your task is to predict what answer the same model would give to the follow-up question. You should base your prediction on the model's answer style, content, and reasoning reflected in its starter answer, even if the starter answer is incomplete or incorrect. Return a short answer only. If the follow-up question is not answerable, return NA.\\
Assistant: Okay.
\end{tcolorbox}

For Phase 1, we use the following template:

\begin{tcolorbox}[
    colback=gray!5,
    colframe=gray!60,
    breakable,
    title=Phase 1 Judge Prompt,
    boxsep=2pt,
    left=4pt,
    right=4pt,
    top=4pt,
    bottom=4pt
]
\small
\setlength{\parskip}{2pt}
Human: Starter Question: \{original\_question\}\\
Model's Answer to the Starter Question: \{model\_answer\}\\
Follow-up Question: \{followup\_question\}\\
Predict the model's answer to the follow-up question.\\
Assistant:
\end{tcolorbox}

For Phase 2, we add the explanation field:

\begin{tcolorbox}[
    colback=gray!5,
    colframe=gray!60,
    breakable,
    title=Phase 2 Judge Prompt,
    boxsep=2pt,
    left=4pt,
    right=4pt,
    top=4pt,
    bottom=4pt
]
\small
\setlength{\parskip}{2pt}
Human: Starter Question: \{original\_question\}\\
Model's Answer to the Starter Question: \{model\_answer\}\\
Model's Explanation to the Starter Question: \{explanation\}\\
Follow-up Question: \{followup\_question\}\\
Predict the model's answer to the follow-up question.\\
Assistant:
\end{tcolorbox}

To make the output parseable, both phases append the same formatting instruction:

\begin{tcolorbox}[
    colback=gray!5,
    colframe=gray!60,
    breakable,
    title=Judge Output Format Instruction,
    boxsep=2pt,
    left=4pt,
    right=4pt,
    top=4pt,
    bottom=4pt
]
\small
\setlength{\parskip}{2pt}
[Instruction] Answer briefly in natural language and end with exactly one non-empty line in this format:\\
PREDICTION: <answer text>\\
If the follow-up question is not answerable, use exactly:\\
PREDICTION: NA\\
Do not leave the prediction blank.
\end{tcolorbox}

\paragraph{Implementation Details.}
We run the judge with deterministic decoding using temperature $0$. The default OpenAI-compatible backend uses \texttt{GPT-5-mini} with a maximum generation length of 512 tokens. The same experimental setting is used for the robustness evaluation with \texttt{DeepSeek-R1}.

The judge output is parsed by extracting the first non-empty line after the string \texttt{PREDICTION:}. We normalize common no-answer strings such as \texttt{NA}, \texttt{N/A}, \texttt{not answerable}, \texttt{unanswerable}, \texttt{cannot answer}, and \texttt{no answer} to a single no-answer label. If no parseable prediction is returned, the output is marked as \texttt{[EMPTY\_PREDICTION]}.

\section{Sample-level Shift Counts}
\label{app:sample_level_shift_counts}

\begin{figure*}[ht]
  \includegraphics[width=\textwidth]{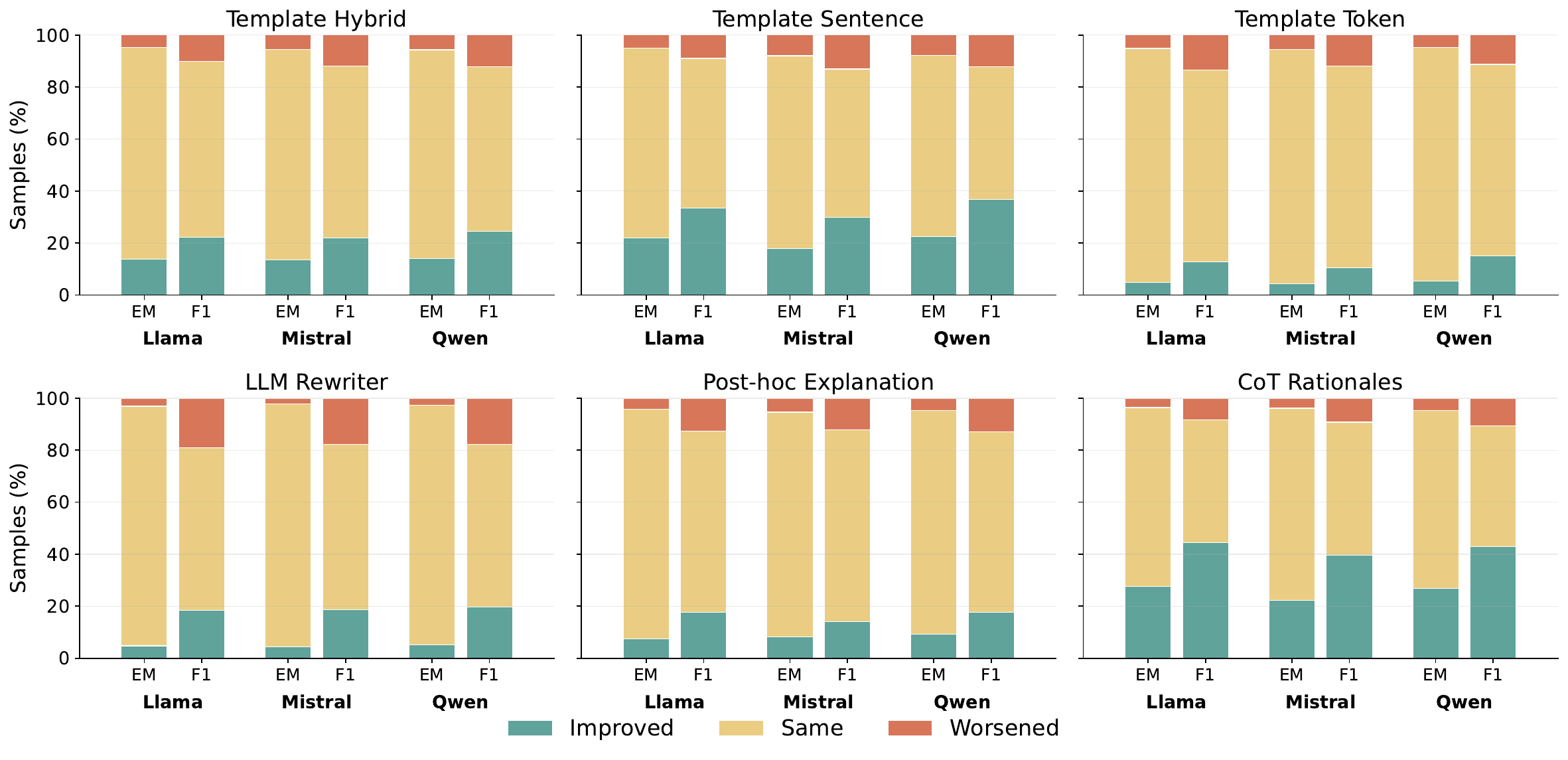}
  \caption{
    Sample-level shifts in counterfactual simulation accuracy on all explanation types. Each subplot corresponds to one explanation condition, and each target model is represented by two stacked bars for \textbf{Exact Match (EM)} and \textbf{F1}.
    }
  \label{fig:sample-level-shifts-full}
\end{figure*}

Figure~\ref{fig:sample-level-shifts-full} presents the complete sample-level shift results across all explanation types, while Table~\ref{tab:sample_shift_counts} reports the underlying raw counts used to construct the figure. Each cell indicates the number of examples for which adding the explanation improves, preserves, or worsens the judge model’s prediction relative to the no-explanation baseline.

Figure~\ref{fig:sample-level-shifts-full} further shows that, CoT rationales produce the largest improved segments across models, especially under F1, indicating that step-by-step rationales often help the simulated user recover partially or fully correct answers. Beyond CoT explanations, \texttt{template\_sentence} exhibits a consistent positive shift pattern: across all three models, the number of improved predictions exceeds the number of worsened predictions. This finding aligns with the aggregate results, suggesting that sentence-level attributions provide particularly useful signals for simulating model behavior. In contrast, \texttt{template\_token} and LLM Rewriter produce much smaller shifts. For \texttt{template\_token}, the improved and worsened segments are relatively balanced, which explains why its aggregate gains are close to zero or negative. LLM Rewriter leaves most examples unchanged and produces only modest differences between improved and worsened cases. Post-hoc explanations fall between: they improve more examples than they worsen, but the shift is much smaller than for CoT rationales or \texttt{template\_sentence}.

Overall, the sample-level analysis reinforces the main finding that explanations differ not only in average \textit{simulatability}, but also in how consistently they help across individual examples. Stronger explanation types produce broad positive shifts, whereas weaker explanation types affect fewer examples and often leave the prediction unchanged.

\section{Additional \texttt{DeepSeek-R1} Judge Evaluation}
\label{app:r1-judge}

\subsection{Robustness Check with \texttt{DeepSeek-R1} as Judge}

As a robustness check, we evaluate the six Llama explanation settings using \texttt{DeepSeek-R1} as an additional LLM-as-judge. We use the same evaluation protocol as in the main results. Table~\ref{tab:r1-judge-results} reports EM and F1 before and after adding explanations.

\begin{table}[h]
\centering
\resizebox{\columnwidth}{!}{
\begin{tabular}{llrr}
\toprule
Model & Explanation &  $\Delta$EM & $\Delta$F1 \\
\midrule
Llama & Template Hybrid   & 3.892  & 3.943  \\
Llama & Template Sentence  & 3.904  & 4.089  \\
Llama & Template Token    & -0.083 & -0.691 \\
Llama & LLM Rewriter   & 0.581  & 0.921  \\
Llama & Post-hoc Expl.  & 1.578  & 1.833  \\
Llama & CoT Rationale     &  17.940 & 22.519 \\
\bottomrule
\end{tabular}
}
\caption{\texttt{DeepSeek-R1} judge evaluation on Llama explanation settings.}
\label{tab:r1-judge-results}
\end{table}

The \texttt{DeepSeek-R1} evaluation broadly supports the main trend. CoT produces by far the largest improvement, increasing EM by $17.94$ and F1 by $22.52$. Among feature-attribution explanations, sentence-level and hybrid explanations provide modest gains, while token-only explanations slightly decrease performance. Rewrite and post-hoc explanations provide only small improvements under \texttt{DeepSeek-R1}, suggesting that their benefit is less robust than CoT or sentence-level attribution evidence.

\subsection{Inter-Judge Agreement}
\label{app:inter-judge-agreement}

We also compare \texttt{DeepSeek-R1} against our main judge \texttt{GPT-5-mini} on the same 1,204 Llama examples. Because the judge output is free-form text, exact agreement is a strict metric; we therefore also report pairwise token-level F1 between judge predictions and agreement on the binary \texttt{NA}/non-\texttt{NA} decision. Table~\ref{tab:inter-judge-agreement} summarizes the Phase 2 agreement, which is the explanation-conditioned setting used for the main comparison.

\begin{table}[h]
\centering
\resizebox{\columnwidth}{!}{
\begin{tabular}{lrrr}
\toprule
Explanation & Exact Agree. & Pairwise F1 & NA Agree. \\
\midrule
Template Hybrid & 35.88 & 39.01 & 43.85 \\
Template Sentence & 28.57 & 32.51 & 40.28 \\
Template Token    & 31.89 & 35.21 & 42.77 \\
LLM Rewriter  & 14.20 & 14.68 & 16.86 \\
Post-hoc Expl. & 35.22 & 39.71 & 48.26 \\
CoT Rationale    & 52.74 & 62.95 & 77.41 \\
\bottomrule
\end{tabular}
}
\caption{Phase 2 inter-judge agreement between \texttt{DeepSeek-R1} and the main judge on Llama explanations. Exact agreement and pairwise F1 compare the free-form predictions, while NA agreement compares the binary answerability decision.}
\label{tab:inter-judge-agreement}
\end{table}

The strongest inter-judge agreement occurs for CoT, where the two judges reach 52.74\% exact agreement, 62.95\% pairwise F1, and 77.41\% agreement on the \texttt{NA}/non-\texttt{NA} decision. Agreement is lower for feature-attribution explanations, reflecting the difficulty of matching free-form counterfactual predictions across judges. The rewrite condition shows particularly weak agreement, largely because R1 predicts \texttt{NA} much more often than the main judge. Overall, the agreement analysis suggests that the main qualitative conclusion is robust for CoT, while judge behavior is more sensitive for shorter or rewritten feature-attribution explanations.

\section{Comparison to Random Selection}
\label{sec:comparision-random-selection}
\begin{figure}[h]
  \includegraphics[width=\columnwidth]{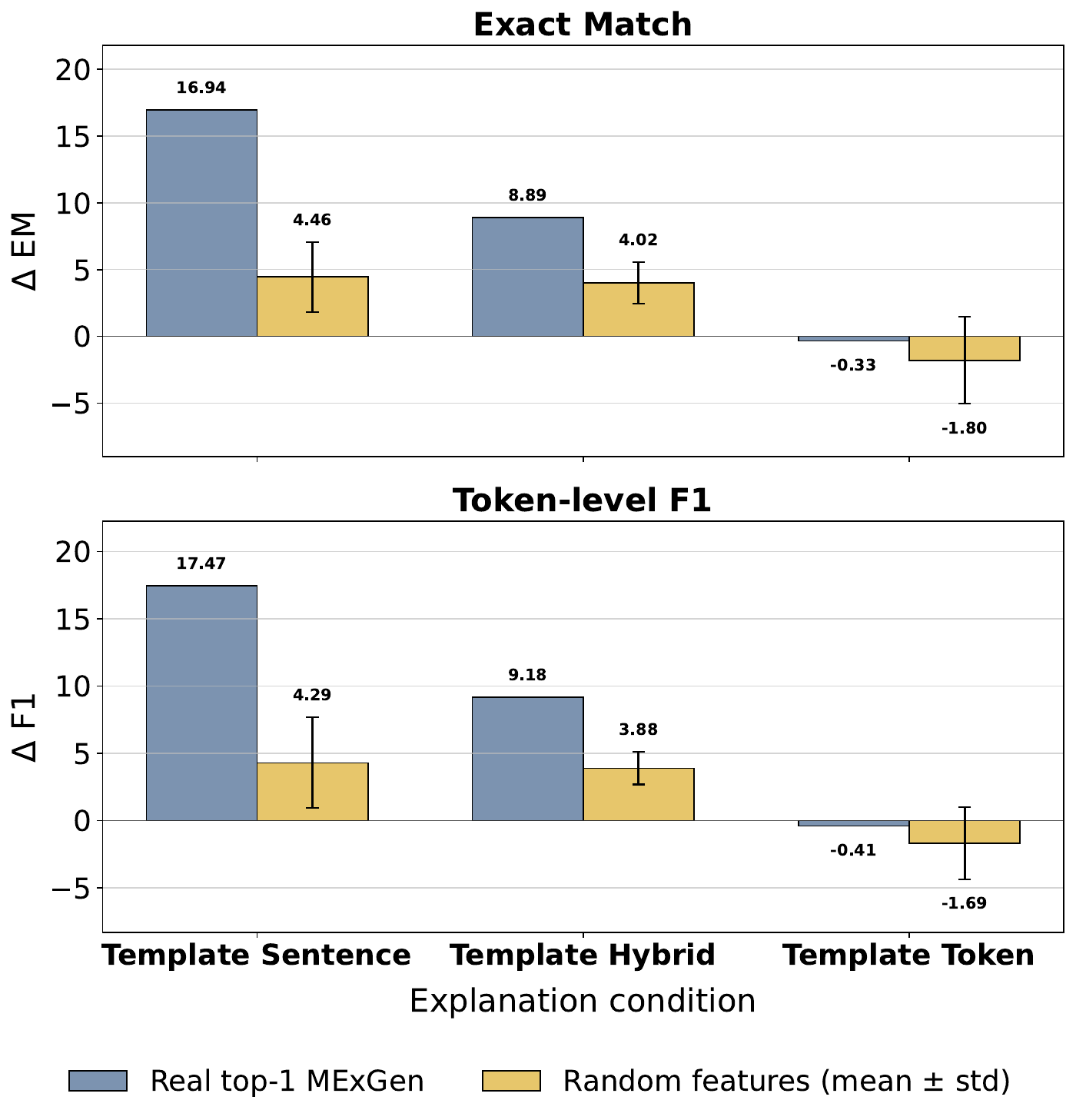}
  \caption{
    Random feature baseline for Llama. Real top-1 MExGen features are compared with randomly selected features across templates. Random bars show the mean and standard deviation over three seeds.
    }
  \label{fig:comparison-random}
\end{figure}

Figure~\ref{fig:comparison-random} compares the top-1 MExGen features with randomly selected features for \texttt{Llama-3}. The random baseline controls for whether the improvement comes from providing any input feature, or from the feature identified as important by MExGen.

The results show that real sentence-level attributions outperform random sentence selection. For \texttt{template\_sentence}, the top-1 MExGen sentence achieves $+16.944$ EM and $+17.474$ F1, while random sentences average only $+4.464$ EM and $+4.289$ F1 across three seeds. This suggests that the top-ranked sentence carries attribution-specific information that supports more accurate prediction of the model's behavior.

A similar pattern appears for \texttt{template\_hybrid}. The real hybrid explanation reaches $+8.887$ EM and $+9.181$ F1, compared with random hybrid averages of $+4.021$ EM and $+3.882$ F1. To conclude, even when random evidence sometimes helps, MExGen-selected evidence provides a stronger and more reliable simulation signal.

For \texttt{template\_token}, both real and random features perform poorly. This reinforces the main result that token-level verbalizations are generally weak for counterfactual simulation, regardless of whether the token is attribution-selected or randomly chosen. Overall, the random baseline supports the idea that the gains from sentence and hybrid explanations are driven by meaningful attribution selection rather than by the mere presence of additional verbalized input features.

\section{Per-model Top-$k$ Ablation Results}
\label{app:topk_ablation_per_model}

Table~\ref{tab:topk_ablation_per_model} reports the per-model results underlying the averaged top-$k$ ablation in Figure~\ref{fig:topk-ablation}. Each entry is reported as Top-1 / Top-2 / Top-3.

\begin{table*}[h]
\centering
\small
\resizebox{0.85\textwidth}{!}{
\begin{tabular}{lccc}
\toprule
Model & Template Sentence & Template Hybrid & Template Token \\
\midrule
\multicolumn{4}{c}{\textbf{$\Delta$ EM: Top-1 / Top-2 / Top-3}} \\
\midrule
Llama-3-8B-Instruct
& 16.944 / 15.356 / 17.691
& 8.887 / 7.226 / 8.056
& -0.332 / 0.997 / 0.083 \\
Mistral-7B-Instruct
& 9.884 / 14.701 / 16.445
& 8.057 / 7.641 / 8.555
& -1.080 / 0.415 / 1.827 \\
Qwen2.5-7B-Instruct-v0.3
& 14.701 / 12.458 / 16.445
& 8.839 / 7.475 / 7.807
& 0.581 / 0.748 / 0.003 \\
\midrule
\multicolumn{4}{c}{\textbf{$\Delta$ F1: Top-1 / Top-2 / Top-3}} \\
\midrule
Llama-3-8B-Instruct
& 17.474 / 17.011 / 19.740
& 9.181 / 7.700 / 7.862
& -0.411 / 0.680 / -0.189 \\
Mistral-7B-Instruct
& 11.999 / 15.977 / 19.492
& 7.999 / 8.565 / 8.523
& -1.372 / 0.024 / 2.301 \\
Qwen2.5-7B-Instruct-v0.3
& 16.296 / 14.052 / 17.967
& 9.343 / 7.948 / 7.779
& 1.382 / 0.367 / 3.301 \\
\bottomrule
\end{tabular}
}
\caption{
Per-model top-$k$ ablation results for template-based verbalized attribution explanations. Each entry is reported as Top-1 / Top-2 / Top-3.
}
\label{tab:topk_ablation_per_model}
\end{table*}

\section{Cross-Model CoT Transfer Results}
\label{sec:cross-model-results}

Table~\ref{tab:cross_model} reports the full $\Delta EM$ and $\Delta F1$ results for all within- and cross-model CoT transfer conditions. Diagonal entries correspond to the standard within-model setting; off-diagonal entries are cross-model transfers.

\begin{table}[h]
\centering
\resizebox{\columnwidth}{!}{%
\begin{tabular}{llcc}
\toprule
Source of CoT & Target Model & $\Delta$EM & $\Delta$F1 \\
\midrule
\multirow{3}{*}{Llama-3-8B-Instruct}
  & Llama (within) & 24.003 & 26.699 \\
  & Mistral        & 16.279 & 18.617 \\
  & Qwen           & 20.086 & 21.021 \\
\midrule
\multirow{3}{*}{Mistral-7B-Instruct}
  & Mistral (within) & 18.355 & 22.374 \\
  & Llama            & 19.020 & 21.713 \\
  & Qwen             & 21.844 & 24.855 \\
\midrule
\multirow{3}{*}{Qwen2.5-7B-Instruct-v0.3}
  & Qwen (within) & 22.259 & 24.024 \\
  & Llama         & 22.841 & 25.086 \\
  & Mistral       & 17.442 & 19.192 \\
\bottomrule
\end{tabular}
}
\caption{Within- and cross-model CoT transfer results. Diagonal entries (within) use the same model's CoT to predict its own counterfactual answers; off-diagonal
entries transfer CoT across models.}
\label{tab:cross_model}
\end{table}

\section{Teacher--Student Evaluation Setups and Results}
\label{sec:teacher-student-setup}

We evaluate explanation quality through a teacher--student simulatability framework. 
Given an input example $x_i=(c_i,q_i)$, where $c_i$ is the context and $q_i$ is the question, we first query a teacher model $f_T$ to obtain the teacher answer $a_i^T=f_T(x_i)$. 
An explanation $e_i^T$ is then constructed as an auxiliary supervision signal that describes the evidence associated with the teacher prediction.

In both settings, the student is required to generate an answer together with a reasoning field, i.e., the output is serialized as $(a_i^T, e_i^T)$. The key difference lies in the supervision signal: the answer-supervised student $S_{\mathrm{ans}}$ is only optimized to imitate the teacher answer, while the explanation-supervised student $S_{\mathrm{exp}}$ is optimized to imitate both the teacher answer and the accompanying explanation. Thus, both students produce explanations at inference time, but only $S_{\mathrm{exp}}$ receives explicit token-level supervision on the explanation field.

For all experiments, we instantiate the student model with \texttt{Qwen2.5-3B-Instruct} and fine-tune it using LoRA, with rank $r=16$, scaling factor $\alpha=32$, and dropout rate $0.05$. 
We train each student for 2 epochs with a learning rate of $2\times10^{-4}$, a per-device batch size of 1, and gradient accumulation over 16 steps.

For the answer-supervised student, we apply a loss mask so that the training objective is computed only over answer tokens:
\begin{equation}
    \mathcal{L}_{\mathrm{ans}}
    =
    -\sum_{t \in \mathcal{A}_i}
    \log p_{\theta}(y_{it}\mid x_i,y_{i,<t}),
\end{equation}
where $\mathcal{A}_i$ denotes the token positions corresponding to the answer field. 
For the explanation-supervised student, we additionally unmask the explanation tokens and compute
\begin{equation}
    \mathcal{L}_{\mathrm{exp}}
    =
    -\sum_{t \in \mathcal{E}_i}
    \log p_{\theta}(y_{it}\mid x_i,y_{i,<t}),
\end{equation}
where $\mathcal{E}_i$ denotes the token positions corresponding to the reasoning field. 
The overall objective is then
\begin{equation}
    \mathcal{L}
    =
    \mathcal{L}_{\mathrm{ans}}
    +
    \lambda \mathcal{L}_{\mathrm{exp}},
\end{equation}
where $\lambda=0$ recovers the answer-supervised setting and $\lambda>0$ enables explicit explanation supervision. 
This design isolates the effect of explanation supervision while keeping the output format and decoding requirement identical across the two student models.

At evaluation time, students are tested on counterfactual inputs $x_i^{\mathrm{cf}}$. 
The student receives only the counterfactual question and context, without access to the teacher explanation. 
We compare the student answer $\hat{a}_i^{S,\mathrm{cf}}$ with the teacher counterfactual answer $a_i^{T,\mathrm{cf}}=f_T(x_i^{\mathrm{cf}})$ using exact match and token-level F1. 
The simulatability gain is defined as $\Delta_{\mathrm{exp}}=\mathrm{Score}(S_{\mathrm{exp}},f_T)-\mathrm{Score}(S_{\mathrm{ans}},f_T)$. 
A positive gain indicates that explanation supervision helps the student better approximate the teacher's behavior on unseen counterfactual examples.


\begin{table*}[ht]
\centering
\small
\resizebox{0.8\textwidth}{!}{%
\begin{tabular}{lcccccc}
\toprule
\multirow{2}{*}{Model}
& \multicolumn{2}{c}{Template Hybrid}
& \multicolumn{2}{c}{Template Sentence}
& \multicolumn{2}{c}{Template Token} \\
\cmidrule(lr){2-3}
\cmidrule(lr){4-5}
\cmidrule(lr){6-7}
& $\Delta EM$ & $\Delta F1$
& $\Delta EM$ & $\Delta F1$
& $\Delta EM$ & $\Delta F1$ \\
\midrule
Llama-3-8B-Instruct  & -0.996 &-2.198  & -0.997 & -2.099 & -0.831 & -1.886 \\
Mistral-7B-Instruct & 0.0831 & -1.193 & 0.000 & -1.083 & 0.000 & -1.419 \\
Qwen2.5-7B-Instruct-v0.3 & 0.000 & 0.000 & 0.166 & 0.199 & 0.000 & 0.270  \\
\midrule
\multirow{2}{*}{Model}
& \multicolumn{2}{c}{LLM Rewriter}
& \multicolumn{2}{c}{Post-hoc Expl.}
& \multicolumn{2}{c}{CoT Rationale} \\
\cmidrule(lr){2-3}
\cmidrule(lr){4-5}
\cmidrule(lr){6-7}
& $\Delta EM$ & $\Delta F1$
& $\Delta EM$ & $\Delta F1$
& $\Delta EM$ & $\Delta F1$ \\
\midrule
Llama-3-8B-Instruct & -0.748 & -0.582 & 0.498 & 0.460 & 15.282 & 14.219 \\
Mistral-7B-Instruct  & 0.000 & 0.095 & 4.319 & 4.080 & 3.312 & 2.474 \\
Qwen2.5-7B-Instruct-v0.3 & 1.329 & 1.048 & 0.830 & 0.622 &9.468 & 8.962 \\
\bottomrule
\end{tabular}
}
\caption{Teacher--Student simulatability results. $\Delta EM$ and $\Delta F1$ report the improvement of the explanation-supervised student over the answer-supervised student on counterfactual inputs. Positive values indicate that explanation supervision helps the student better approximate the teacher's counterfactual behavior.
}
\label{tab:teacher-student-results}
\end{table*}

%% file: custom.bib
@article{Agarwal2024FaithfulnessVP,
  title={Faithfulness vs. Plausibility: On the (Un)Reliability of Explanations from Large Language Models},
  author={Chirag Agarwal and Sree Harsha Tanneru and Himabindu Lakkaraju},
  journal={ArXiv},
  year={2024},
  volume={abs/2402.04614},
  url={https://api.semanticscholar.org/CorpusID:267523276}
}

@inproceedings{atanasova-etal-2023-faithfulness,
    title = "Faithfulness Tests for Natural Language Explanations",
    author = "Atanasova, Pepa  and
      Camburu, Oana-Maria  and
      Lioma, Christina  and
      Lukasiewicz, Thomas  and
      Simonsen, Jakob Grue  and
      Augenstein, Isabelle",
    editor = "Rogers, Anna  and
      Boyd-Graber, Jordan  and
      Okazaki, Naoaki",
    booktitle = "Proceedings of the 61st Annual Meeting of the Association for Computational Linguistics (Volume 2: Short Papers)",
    month = jul,
    year = "2023",
    address = "Toronto, Canada",
    publisher = "Association for Computational Linguistics",
    url = "https://aclanthology.org/2023.acl-short.25/",
    doi = "10.18653/v1/2023.acl-short.25",
    pages = "283--294"
}

@article{Bai2023BenchmarkingFM,
  title={Benchmarking Foundation Models with Language-Model-as-an-Examiner},
  author={Yushi Bai and Jiahao Ying and Yixin Cao and Xin Lv and Yuze He and Xiaozhi Wang and Jifan Yu and Kaisheng Zeng and Yijia Xiao and Haozhe Lyu and Jiayin Zhang and Juanzi Li and Lei Hou},
  journal={ArXiv},
  year={2023},
  volume={abs/2306.04181},
  url={https://api.semanticscholar.org/CorpusID:259095491}
}

@article{bansal_beyond_2019,
	title = {Beyond Accuracy: The Role of Mental Models in Human-{AI} Team Performance},
	volume = {7},
    year ={2019},
	url = {https://ojs.aaai.org/index.php/HCOMP/article/view/5285},
	doi = {10.1609/hcomp.v7i1.5285},
	pages = {2--11},
	number = {1},
	journaltitle = {Proceedings of the {AAAI} Conference on Human Computation and Crowdsourcing},
	author = {Bansal, Gagan and Nushi, Besmira and Kamar, Ece and Lasecki, Walter S. and Weld, Daniel S. and Horvitz, Eric},
	date = {2019-10},
}

@inproceedings{boyd-graber-etal-2022-human,
    title = "Human-Centered Evaluation of Explanations",
    author = "Boyd-Graber, Jordan  and
      Carton, Samuel  and
      Feng, Shi  and
      Liao, Q. Vera  and
      Lombrozo, Tania  and
      Smith-Renner, Alison  and
      Tan, Chenhao",
    editor = "Ballesteros, Miguel  and
      Tsvetkov, Yulia  and
      Alm, Cecilia O.",
    booktitle = "Proceedings of the 2022 Conference of the North American Chapter of the Association for Computational Linguistics: Human Language Technologies: Tutorial Abstracts",
    month = jul,
    year = "2022",
    address = "Seattle, United States",
    publisher = "Association for Computational Linguistics",
    url = "https://aclanthology.org/2022.naacl-tutorials.4/",
    doi = "10.18653/v1/2022.naacl-tutorials.4",
    pages = "26--32"
}

@misc{brandl2026systematiccomparisonextractiveselfexplanations,
      title={A Systematic Comparison between Extractive Self-Explanations and Human Rationales in Text Classification}, 
      author={Stephanie Brandl and Oliver Eberle},
      year={2026},
      eprint={2410.03296},
      archivePrefix={arXiv},
      primaryClass={cs.CL},
      url={https://arxiv.org/abs/2410.03296}, 
}

@inproceedings{Camburu2018eSNLINL,
  title={e-{SNLI}: Natural Language Inference with Natural Language Explanations},
  author={Oana-Maria Camburu and Tim Rockt{\"a}schel and Thomas Lukasiewicz and Phil Blunsom},
  booktitle={Neural Information Processing Systems},
  year={2018},
  url={https://api.semanticscholar.org/CorpusID:54040953}
}

@inproceedings{chandrasekaran-etal-2018-explanations,
    title = "Do explanations make {VQA} models more predictable to a human?",
    author = "Chandrasekaran, Arjun  and
      Prabhu, Viraj  and
      Yadav, Deshraj  and
      Chattopadhyay, Prithvijit  and
      Parikh, Devi",
    editor = "Riloff, Ellen  and
      Chiang, David  and
      Hockenmaier, Julia  and
      Tsujii, Jun{'}ichi",
    booktitle = "Proceedings of the 2018 Conference on Empirical Methods in Natural Language Processing",
    month = oct # "-" # nov,
    year = "2018",
    address = "Brussels, Belgium",
    publisher = "Association for Computational Linguistics",
    url = "https://aclanthology.org/D18-1128/",
    doi = "10.18653/v1/D18-1128",
    pages = "1036--1042"
}

@misc{chen2023modelsexplainthemselvescounterfactual,
      title={Do Models Explain Themselves? Counterfactual Simulatability of Natural Language Explanations}, 
      author={Yanda Chen and Ruiqi Zhong and Narutatsu Ri and Chen Zhao and He He and Jacob Steinhardt and Zhou Yu and Kathleen McKeown},
      year={2023},
      eprint={2307.08678},
      archivePrefix={arXiv},
      primaryClass={cs.CL},
      url={https://arxiv.org/abs/2307.08678}, 
}

@misc{doshivelez2017towards,
  author = {Doshi-Velez, Finale and Kim, Been},
  description = {Towards A Rigorous Science of Interpretable Machine Learning},
  note = {cite arxiv:1702.08608},
  title = {Towards A Rigorous Science of Interpretable Machine Learning},
  url = {http://arxiv.org/abs/1702.08608},
  year = 2017
}

@inproceedings{enouen-etal-2024-textgenshap,
    title = "{T}ext{G}en{SHAP}: Scalable Post-Hoc Explanations in Text Generation with Long Documents",
    author = "Enouen, James  and
      Nakhost, Hootan  and
      Ebrahimi, Sayna  and
      Arik, Sercan  and
      Liu, Yan  and
      Pfister, Tomas",
    editor = "Ku, Lun-Wei  and
      Martins, Andre  and
      Srikumar, Vivek",
    booktitle = "Findings of the Association for Computational Linguistics: ACL 2024",
    month = aug,
    year = "2024",
    address = "Bangkok, Thailand",
    publisher = "Association for Computational Linguistics",
    url = "https://aclanthology.org/2024.findings-acl.832/",
    doi = "10.18653/v1/2024.findings-acl.832",
    pages = "13984--14011"
}

@inproceedings{feldhus-etal-2023-saliency,
    title = "Saliency Map Verbalization: Comparing Feature Importance Representations from Model-free and Instruction-based Methods",
    author = {Feldhus, Nils  and
      Hennig, Leonhard  and
      Nasert, Maximilian Dustin  and
      Ebert, Christopher  and
      Schwarzenberg, Robert  and
      M{\"o}ller, Sebastian},
    editor = "Dalvi Mishra, Bhavana  and
      Durrett, Greg  and
      Jansen, Peter  and
      Neves Ribeiro, Danilo  and
      Wei, Jason",
    booktitle = "Proceedings of the 1st Workshop on Natural Language Reasoning and Structured Explanations (NLRSE)",
    month = jun,
    year = "2023",
    address = "Toronto, Canada",
    publisher = "Association for Computational Linguistics",
    url = "https://aclanthology.org/2023.nlrse-1.4/",
    doi = "10.18653/v1/2023.nlrse-1.4",
    pages = "30--46"
}

@inproceedings{Fragkathoulas_2024, series={SETN 2024},
   title={Local Explanations and Self-Explanations for Assessing Faithfulness in black-box LLMs},
   url={http://dx.doi.org/10.1145/3688671.3688775},
   DOI={10.1145/3688671.3688775},
   booktitle={Proceedings of the 13th Hellenic Conference on Artificial Intelligence},
   publisher={ACM},
   author={Fragkathoulas, Christos and Chlapanis, Odysseas Spyridon},
   year={2024},
   pages={1–5},
   collection={SETN 2024} }

@misc{grattafiori2024llama3herdmodels,
  title={The {L}lama 3 Herd of Models},
  author={Aaron Grattafiori and Abhimanyu Dubey and Abhinav Jauhri and others},
  year={2024},
  eprint={2407.21783},
  archivePrefix={arXiv},
  primaryClass={cs.AI},
  url={https://arxiv.org/abs/2407.21783}
}

@article{Guo_2025,
   title={Deep{S}eek-{R}1 incentivizes reasoning in LLMs through reinforcement learning},
   author={Guo, Daya and Yang, Dejian and Zhang, Haowei and others},
   journal={Nature},
   volume={645},
   number={8081},
   pages={633--638},
   year={2025},
   month={Sep},
   doi={10.1038/s41586-025-09422-z}
}

@inproceedings{hase-bansal-2020-evaluating,
    title = "Evaluating Explainable {AI}: Which Algorithmic Explanations Help Users Predict Model Behavior?",
    author = "Hase, Peter  and
      Bansal, Mohit",
    editor = "Jurafsky, Dan  and
      Chai, Joyce  and
      Schluter, Natalie  and
      Tetreault, Joel",
    booktitle = "Proceedings of the 58th Annual Meeting of the Association for Computational Linguistics",
    month = jul,
    year = "2020",
    address = "Online",
    publisher = "Association for Computational Linguistics",
    url = "https://aclanthology.org/2020.acl-main.491/",
    doi = "10.18653/v1/2020.acl-main.491",
    pages = "5540--5552"
}

@inproceedings{hase-etal-2020-leakage,
    title = "Leakage-Adjusted Simulatability: Can Models Generate Non-Trivial Explanations of Their Behavior in Natural Language?",
    author = "Hase, Peter  and
      Zhang, Shiyue  and
      Xie, Harry  and
      Bansal, Mohit",
    editor = "Cohn, Trevor  and
      He, Yulan  and
      Liu, Yang",
    booktitle = "Findings of the Association for Computational Linguistics: EMNLP 2020",
    month = nov,
    year = "2020",
    address = "Online",
    publisher = "Association for Computational Linguistics",
    url = "https://aclanthology.org/2020.findings-emnlp.390/",
    doi = "10.18653/v1/2020.findings-emnlp.390",
    pages = "4351--4367"
}

@article{Hong2026DoLS,
  title={Do {LLM} Self-Explanations Help Users Predict Model Behavior? {E}valuating Counterfactual Simulatability with Pragmatic Perturbations},
  author={Pingjun Hong and Benjamin Roth},
  journal={ArXiv},
  year={2026},
  volume={abs/2601.03775},
  url={https://api.semanticscholar.org/CorpusID:284532077}
}

@article{Huang2023ASO,
  title={A Survey on Hallucination in Large Language Models: Principles, Taxonomy, Challenges, and Open Questions},
  author={Lei Huang and Weijiang Yu and Weitao Ma and Weihong Zhong and Zhangyin Feng and Haotian Wang and Qianglong Chen and Weihua Peng and Xiaocheng Feng and Bing Qin and Ting Liu},
  journal={ACM Transactions on Information Systems},
  year={2023},
  volume={43},
  pages={1 - 55},
  url={https://api.semanticscholar.org/CorpusID:265067168}
}

@misc{huang2023largelanguagemodelsexplain,
      title={Can Large Language Models Explain Themselves? A Study of LLM-Generated Self-Explanations}, 
      author={Shiyuan Huang and Siddarth Mamidanna and Shreedhar Jangam and Yilun Zhou and Leilani H. Gilpin},
      year={2023},
      eprint={2310.11207},
      archivePrefix={arXiv},
      primaryClass={cs.CL},
      url={https://arxiv.org/abs/2310.11207}, 
}

@inproceedings{jacovi-goldberg-2020-towards,
    title = "Towards Faithfully Interpretable {NLP} Systems: How Should We Define and Evaluate Faithfulness?",
    author = "Jacovi, Alon  and
      Goldberg, Yoav",
    editor = "Jurafsky, Dan  and
      Chai, Joyce  and
      Schluter, Natalie  and
      Tetreault, Joel",
    booktitle = "Proceedings of the 58th Annual Meeting of the Association for Computational Linguistics",
    month = jul,
    year = "2020",
    address = "Online",
    publisher = "Association for Computational Linguistics",
    url = "https://aclanthology.org/2020.acl-main.386/",
    doi = "10.18653/v1/2020.acl-main.386",
    pages = "4198--4205"
}

@misc{jiang2023mistral7b,
      title={Mistral 7{B}}, 
      author={Albert Q. Jiang and Alexandre Sablayrolles and Arthur Mensch and Chris Bamford and Devendra Singh Chaplot and Diego de las Casas and Florian Bressand and Gianna Lengyel and Guillaume Lample and Lucile Saulnier and Lélio Renard Lavaud and Marie-Anne Lachaux and Pierre Stock and Teven Le Scao and Thibaut Lavril and Thomas Wang and Timothée Lacroix and William El Sayed},
      year={2023},
      eprint={2310.06825},
      archivePrefix={arXiv},
      primaryClass={cs.CL},
      url={https://arxiv.org/abs/2310.06825}, 
}

@misc{kroeger2024incontextexplainersharnessingllms,
      title={In-Context Explainers: Harnessing LLMs for Explaining Black Box Models}, 
      author={Nicholas Kroeger and Dan Ley and Satyapriya Krishna and Chirag Agarwal and Himabindu Lakkaraju},
      year={2024},
      eprint={2310.05797},
      archivePrefix={arXiv},
      primaryClass={cs.CL},
      url={https://arxiv.org/abs/2310.05797}, 
}

@inproceedings{lei-etal-2016-rationalizing,
    title = "Rationalizing Neural Predictions",
    author = "Lei, Tao  and
      Barzilay, Regina  and
      Jaakkola, Tommi",
    editor = "Su, Jian  and
      Duh, Kevin  and
      Carreras, Xavier",
    booktitle = "Proceedings of the 2016 Conference on Empirical Methods in Natural Language Processing",
    month = nov,
    year = "2016",
    address = "Austin, Texas",
    publisher = "Association for Computational Linguistics",
    url = "https://aclanthology.org/D16-1011/",
    doi = "10.18653/v1/D16-1011",
    pages = "107--117"
}

@inproceedings{li-etal-2025-generation,
    title = "From Generation to Judgment: Opportunities and Challenges of {LLM}-as-a-judge",
    author = "Li, Dawei  and
      Jiang, Bohan  and
      Huang, Liangjie  and
      Beigi, Alimohammad  and
      Zhao, Chengshuai  and
      Tan, Zhen  and
      Bhattacharjee, Amrita  and
      Jiang, Yuxuan  and
      Chen, Canyu  and
      Wu, Tianhao  and
      Shu, Kai  and
      Cheng, Lu  and
      Liu, Huan",
    editor = "Christodoulopoulos, Christos  and
      Chakraborty, Tanmoy  and
      Rose, Carolyn  and
      Peng, Violet",
    booktitle = "Proceedings of the 2025 Conference on Empirical Methods in Natural Language Processing",
    month = nov,
    year = "2025",
    address = "Suzhou, China",
    publisher = "Association for Computational Linguistics",
    url = "https://aclanthology.org/2025.emnlp-main.138/",
    doi = "10.18653/v1/2025.emnlp-main.138",
    pages = "2757--2791",
    ISBN = "979-8-89176-332-6"
}

@inproceedings{limpijankit-etal-2025-counterfactual,
    title = "Counterfactual Simulatability of {LLM} Explanations for Generation Tasks",
    author = "Limpijankit, Marvin  and
      Chen, Yanda  and
      Subbiah, Melanie  and
      Deas, Nicholas  and
      McKeown, Kathleen",
    editor = "Flek, Lucie  and
      Narayan, Shashi  and
      Phương, L{\^e} Hồng  and
      Pei, Jiahuan",
    booktitle = "Proceedings of the 18th International Natural Language Generation Conference",
    month = oct,
    year = "2025",
    address = "Hanoi, Vietnam",
    publisher = "Association for Computational Linguistics",
    url = "https://aclanthology.org/2025.inlg-main.38/",
    pages = "659--683"
}

@inproceedings{Lundberg2017AUA,
  title={A Unified Approach to Interpreting Model Predictions},
  author={Scott M. Lundberg and Su-In Lee},
  booktitle={Neural Information Processing Systems},
  year={2017},
  url={https://api.semanticscholar.org/CorpusID:21889700}
}

@inproceedings{madsen-etal-2024-self,
    title = "Are self-explanations from Large Language Models faithful?",
    author = "Madsen, Andreas  and
      Chandar, Sarath  and
      Reddy, Siva",
    editor = "Ku, Lun-Wei  and
      Martins, Andre  and
      Srikumar, Vivek",
    booktitle = "Findings of the Association for Computational Linguistics: ACL 2024",
    month = aug,
    year = "2024",
    address = "Bangkok, Thailand",
    publisher = "Association for Computational Linguistics",
    url = "https://aclanthology.org/2024.findings-acl.19/",
    doi = "10.18653/v1/2024.findings-acl.19",
    pages = "295--337"
}

@misc{mayne2026positivecasefaithfulnessllm,
      title={A Positive Case for Faithfulness: {LLM} Self-Explanations Help Predict Model Behavior}, 
      author={Harry Mayne and Justin Singh Kang and Dewi Gould and Kannan Ramchandran and Adam Mahdi and Noah Y. Siegel},
      year={2026},
      eprint={2602.02639},
      archivePrefix={arXiv},
      primaryClass={cs.AI},
      url={https://arxiv.org/abs/2602.02639}, 
}

@misc{miller_explanation_2018,
	title = {Explanation in Artificial Intelligence: Insights from the Social Sciences},
	url = {http://arxiv.org/abs/1706.07269},
	doi = {10.48550/arXiv.1706.07269},
	shorttitle = {Explanation in Artificial Intelligence},
	number = {{arXiv}:1706.07269},
	publisher = {{arXiv}},
	author = {Miller, Tim},
	urldate = {2025-09-02},
	date = {2018-08-15},
    year = {2018},
	langid = {american},
	eprinttype = {arxiv},
	eprint = {1706.07269 [cs]},
}

@inproceedings{monteiro-paes-etal-2025-multi,
    title = "Multi-Level Explanations for Generative Language Models",
    author = "Monteiro Paes, Lucas  and
      Wei, Dennis  and
      Do, Hyo Jin  and
      Strobelt, Hendrik  and
      Luss, Ronny  and
      Dhurandhar, Amit  and
      Nagireddy, Manish  and
      Natesan Ramamurthy, Karthikeyan  and
      Sattigeri, Prasanna  and
      Geyer, Werner  and
      Ghosh, Soumya",
    editor = "Che, Wanxiang  and
      Nabende, Joyce  and
      Shutova, Ekaterina  and
      Pilehvar, Mohammad Taher",
    booktitle = "Proceedings of the 63rd Annual Meeting of the Association for Computational Linguistics (Volume 1: Long Papers)",
    month = jul,
    year = "2025",
    address = "Vienna, Austria",
    publisher = "Association for Computational Linguistics",
    url = "https://aclanthology.org/2025.acl-long.1553/",
    doi = "10.18653/v1/2025.acl-long.1553",
    pages = "32291--32317",
    ISBN = "979-8-89176-251-0"
}

@inproceedings{mosca-etal-2022-shap,
    title = "{SHAP}-Based Explanation Methods: A Review for {NLP} Interpretability",
    author = "Mosca, Edoardo  and
      Szigeti, Ferenc  and
      Tragianni, Stella  and
      Gallagher, Daniel  and
      Groh, Georg",
    editor = "Calzolari, Nicoletta  and
      Huang, Chu-Ren  and
      Kim, Hansaem  and
      Pustejovsky, James  and
      Wanner, Leo  and
      Choi, Key-Sun  and
      Ryu, Pum-Mo  and
      Chen, Hsin-Hsi  and
      Donatelli, Lucia  and
      Ji, Heng  and
      Kurohashi, Sadao  and
      Paggio, Patrizia  and
      Xue, Nianwen  and
      Kim, Seokhwan  and
      Hahm, Younggyun  and
      He, Zhong  and
      Lee, Tony Kyungil  and
      Santus, Enrico  and
      Bond, Francis  and
      Na, Seung-Hoon",
    booktitle = "Proceedings of the 29th International Conference on Computational Linguistics",
    month = oct,
    year = "2022",
    address = "Gyeongju, Republic of Korea",
    publisher = "International Committee on Computational Linguistics",
    url = "https://aclanthology.org/2022.coling-1.406/",
    pages = "4593--4603"
}

@inproceedings{nguyen-2018-comparing,
    title = "Comparing Automatic and Human Evaluation of Local Explanations for Text Classification",
    author = "Nguyen, Dong",
    editor = "Walker, Marilyn  and
      Ji, Heng  and
      Stent, Amanda",
    booktitle = "Proceedings of the 2018 Conference of the North {A}merican Chapter of the Association for Computational Linguistics: Human Language Technologies, Volume 1 (Long Papers)",
    month = jun,
    year = "2018",
    address = "New Orleans, Louisiana",
    publisher = "Association for Computational Linguistics",
    url = "https://aclanthology.org/N18-1097/",
    doi = "10.18653/v1/N18-1097",
    pages = "1069--1078"
}

@article{Nye2021ShowYW,
  title={Show Your Work: Scratchpads for Intermediate Computation with Language Models},
  author={Maxwell Nye and Anders Andreassen and Guy Gur-Ari and Henryk Michalewski and Jacob Austin and David Bieber and David Dohan and Aitor Lewkowycz and Maarten Bosma and David Luan and Charles Sutton and Augustus Odena},
  journal={ArXiv},
  year={2021},
  volume={abs/2112.00114},
  url={https://api.semanticscholar.org/CorpusID:244773644}
}

@article{Park2018MultimodalEJ,
  title={Multimodal Explanations: Justifying Decisions and Pointing to the Evidence},
  author={Dong Huk Park and Lisa Anne Hendricks and Zeynep Akata and Anna Rohrbach and Bernt Schiele and Trevor Darrell and Marcus Rohrbach},
  journal={2018 IEEE/CVF Conference on Computer Vision and Pattern Recognition},
  year={2018},
  pages={8779-8788},
  url={https://api.semanticscholar.org/CorpusID:3604848}
}

@inproceedings{poche-etal-2025-consim,
    title = "{C}on{S}im: Measuring Concept-Based Explanations' Effectiveness with Automated Simulatability",
    author = "Poch{\'e}, Antonin  and
      Jacovi, Alon  and
      Picard, Agustin Martin  and
      Boutin, Victor  and
      Jourdan, Fanny",
    editor = "Che, Wanxiang  and
      Nabende, Joyce  and
      Shutova, Ekaterina  and
      Pilehvar, Mohammad Taher",
    booktitle = "Proceedings of the 63rd Annual Meeting of the Association for Computational Linguistics (Volume 1: Long Papers)",
    month = jul,
    year = "2025",
    address = "Vienna, Austria",
    publisher = "Association for Computational Linguistics",
    url = "https://aclanthology.org/2025.acl-long.279/",
    doi = "10.18653/v1/2025.acl-long.279",
    pages = "5594--5615",
    ISBN = "979-8-89176-251-0"
}

@article{pruthi_evaluating_2022,
	title = {Evaluating Explanations: How Much Do Explanations from the Teacher Aid Students?},
    year = {2022},
	volume = {10},
	issn = {2307-387X},
	url = {https://direct.mit.edu/tacl/article/doi/10.1162/tacl_a_00465/110436/Evaluating-Explanations-How-Much-Do-Explanations},
	doi = {10.1162/tacl_a_00465},
	shorttitle = {Evaluating Explanations},
	pages = {359--375},
	journaltitle = {Transactions of the Association for Computational Linguistics},
	author = {Pruthi, Danish and Bansal, Rachit and Dhingra, Bhuwan and Soares, Livio Baldini and Collins, Michael and Lipton, Zachary C. and Neubig, Graham and Cohen, William W.},
	urldate = {2026-04-22},
	date = {2022-04-06},
	langid = {english},
}

@misc{qwen2025qwen25technicalreport,
      title={Qwen2.5 Technical Report}, 
      author={Qwen and An Yang and Baosong Yang and Beichen Zhang and Binyuan Hui and Bo Zheng and Bowen Yu and Chengyuan Li and Dayiheng Liu and Fei Huang and Haoran Wei and Huan Lin and Jian Yang and Jianhong Tu and Jianwei Zhang and Jianxin Yang and Jiaxi Yang and Jingren Zhou and Junyang Lin and Kai Dang and Keming Lu and Keqin Bao and Kexin Yang and Le Yu and Mei Li and Mingfeng Xue and Pei Zhang and Qin Zhu and Rui Men and Runji Lin and Tianhao Li and Tianyi Tang and Tingyu Xia and Xingzhang Ren and Xuancheng Ren and Yang Fan and Yang Su and Yichang Zhang and Yu Wan and Yuqiong Liu and Zeyu Cui and Zhenru Zhang and Zihan Qiu},
      year={2025},
      eprint={2412.15115},
      archivePrefix={arXiv},
      primaryClass={cs.CL},
      url={https://arxiv.org/abs/2412.15115}, 
}

@inproceedings{rajpurkar-etal-2016-squad,
    title = "{SQ}u{AD}: 100,000+ Questions for Machine Comprehension of Text",
    author = "Rajpurkar, Pranav  and
      Zhang, Jian  and
      Lopyrev, Konstantin  and
      Liang, Percy",
    editor = "Su, Jian  and
      Duh, Kevin  and
      Carreras, Xavier",
    booktitle = "Proceedings of the 2016 Conference on Empirical Methods in Natural Language Processing",
    month = nov,
    year = "2016",
    address = "Austin, Texas",
    publisher = "Association for Computational Linguistics",
    url = "https://aclanthology.org/D16-1264",
    doi = "10.18653/v1/D16-1264",
    pages = "2383--2392",
    eprint={1606.05250},
    archivePrefix={arXiv},
    primaryClass={cs.CL},
}

@misc{rajpurkar2018knowdontknowunanswerable,
      title={Know What You Don't Know: Unanswerable Questions for SQuAD}, 
      author={Pranav Rajpurkar and Robin Jia and Percy Liang},
      year={2018},
      eprint={1806.03822},
      archivePrefix={arXiv},
      primaryClass={cs.CL},
      url={https://arxiv.org/abs/1806.03822}, 
}

@inproceedings{ribeiro-etal-2016-trust,
    title = "``{W}hy Should {I} Trust You?'': Explaining the Predictions of Any Classifier",
    author = "Ribeiro, Marco  and
      Singh, Sameer  and
      Guestrin, Carlos",
    editor = "DeNero, John  and
      Finlayson, Mark  and
      Reddy, Sravana",
    booktitle = "Proceedings of the 2016 Conference of the North {A}merican Chapter of the Association for Computational Linguistics: Demonstrations",
    month = jun,
    year = "2016",
    address = "San Diego, California",
    publisher = "Association for Computational Linguistics",
    url = "https://aclanthology.org/N16-3020/",
    doi = "10.18653/v1/N16-3020",
    pages = "97--101"
}

@inproceedings{ross-etal-2021-explaining,
    title = "Explaining {NLP} Models via Minimal Contrastive Editing ({M}i{CE})",
    author = "Ross, Alexis  and
      Marasovi{\'c}, Ana  and
      Peters, Matthew",
    editor = "Zong, Chengqing  and
      Xia, Fei  and
      Li, Wenjie  and
      Navigli, Roberto",
    booktitle = "Findings of the Association for Computational Linguistics: ACL-IJCNLP 2021",
    month = aug,
    year = "2021",
    address = "Online",
    publisher = "Association for Computational Linguistics",
    url = "https://aclanthology.org/2021.findings-acl.336/",
    doi = "10.18653/v1/2021.findings-acl.336",
    pages = "3840--3852"
}

@inproceedings{saphra-wiegreffe-2024-mechanistic,
    title = "Mechanistic?",
    author = "Saphra, Naomi  and
      Wiegreffe, Sarah",
    editor = "Belinkov, Yonatan  and
      Kim, Najoung  and
      Jumelet, Jaap  and
      Mohebbi, Hosein  and
      Mueller, Aaron  and
      Chen, Hanjie",
    booktitle = "Proceedings of the 7th BlackboxNLP Workshop: Analyzing and Interpreting Neural Networks for NLP",
    month = nov,
    year = "2024",
    address = "Miami, Florida, US",
    publisher = "Association for Computational Linguistics",
    url = "https://aclanthology.org/2024.blackboxnlp-1.30/",
    doi = "10.18653/v1/2024.blackboxnlp-1.30",
    pages = "480--498"
}

@article{Shen2025FaithCoTBenchBI,
  title={FaithCoT-Bench: Benchmarking Instance-Level Faithfulness of Chain-of-Thought Reasoning},
  author={Xu Shen and Song Wang and Zhen Tan and Laura Yao and Xinyu Zhao and Kaidi Xu and Xin Wang and Tianlong Chen},
  journal={ArXiv},
  year={2025},
  volume={abs/2510.04040},
  url={https://api.semanticscholar.org/CorpusID:281844153}
}

@misc{singh2025openaigpt5card,
  title={OpenAI GPT-5 System Card},
  author={Aaditya Singh and Adam Fry and Adam Perelman and et al.},
  year={2025},
  eprint={2601.03267},
  archivePrefix={arXiv},
  primaryClass={cs.CL},
  url={https://arxiv.org/abs/2601.03267},
}

@article{Turpin2023LanguageMD,
  title={Language Models Don't Always Say What They Think: Unfaithful Explanations in Chain-of-Thought Prompting},
  author={Miles Turpin and Julian Michael and Ethan Perez and Sam Bowman},
  journal={ArXiv},
  year={2023},
  volume={abs/2305.04388},
  url={https://api.semanticscholar.org/CorpusID:258556812}
}

@inproceedings{cotexplanation2022,
  author       = {Jason Wei and
                  Xuezhi Wang and
                  Dale Schuurmans and
                  Maarten Bosma and
                  Brian Ichter and
                  Fei Xia and
                  Ed H. Chi and
                  Quoc V. Le and
                  Denny Zhou},
  editor       = {Sanmi Koyejo and
                  S. Mohamed and
                  A. Agarwal and
                  Danielle Belgrave and
                  K. Cho and
                  A. Oh},
  title        = {Chain-of-Thought Prompting Elicits Reasoning in Large Language Models},
  booktitle    = {Advances in Neural Information Processing Systems 35: Annual Conference
                  on Neural Information Processing Systems 2022, NeurIPS 2022, New Orleans,
                  LA, USA, November 28 - December 9, 2022},
  year         = {2022},
  url          = {http://papers.nips.cc/paper\_files/paper/2022/hash/9d5609613524ecf4f15af0f7b31abca4-Abstract-Conference.html},
  bibsource    = {dblp computer science bibliography, https://dblp.org}
}

@inproceedings{wei-jie-etal-2024-interpretable,
    title = "How Interpretable are Reasoning Explanations from Prompting Large Language Models?",
    author = "Wei Jie, Yeo  and
      Satapathy, Ranjan  and
      Goh, Rick  and
      Cambria, Erik",
    editor = "Duh, Kevin  and
      Gomez, Helena  and
      Bethard, Steven",
    booktitle = "Findings of the Association for Computational Linguistics: NAACL 2024",
    month = jun,
    year = "2024",
    address = "Mexico City, Mexico",
    publisher = "Association for Computational Linguistics",
    url = "https://aclanthology.org/2024.findings-naacl.138/",
    doi = "10.18653/v1/2024.findings-naacl.138",
    pages = "2148--2164"
}

@misc{wei2025icx360incontextexplainability360,
      title={{I}{C}{X}360: In-Context eXplainability 360 Toolkit}, 
      author={Dennis Wei and Ronny Luss and Xiaomeng Hu and Lucas Monteiro Paes and Pin-Yu Chen and Karthikeyan Natesan Ramamurthy and Erik Miehling and Inge Vejsbjerg and Hendrik Strobelt},
      year={2025},
      eprint={2511.10879},
      archivePrefix={arXiv},
      primaryClass={cs.CL},
      url={https://arxiv.org/abs/2511.10879}, 
}

@inproceedings{Wiegreffe2021TeachMT,
  title={Teach Me to Explain: A Review of Datasets for Explainable Natural Language Processing},
  author={Sarah Wiegreffe and Ana Marasovi{\'c}},
  booktitle={NeurIPS Datasets and Benchmarks},
  year={2021},
  url={https://api.semanticscholar.org/CorpusID:232035689}
}

@inproceedings{wiegreffe-etal-2021-measuring,
    title = "{M}easuring Association Between Labels and Free-Text Rationales",
    author = "Wiegreffe, Sarah  and
      Marasovi{\'c}, Ana  and
      Smith, Noah A.",
    editor = "Moens, Marie-Francine  and
      Huang, Xuanjing  and
      Specia, Lucia  and
      Yih, Scott Wen-tau",
    booktitle = "Proceedings of the 2021 Conference on Empirical Methods in Natural Language Processing",
    month = nov,
    year = "2021",
    address = "Online and Punta Cana, Dominican Republic",
    publisher = "Association for Computational Linguistics",
    url = "https://aclanthology.org/2021.emnlp-main.804/",
    doi = "10.18653/v1/2021.emnlp-main.804",
    pages = "10266--10284"
}

@inproceedings{wu-etal-2021-polyjuice,
    title = "Polyjuice: Generating Counterfactuals for Explaining, Evaluating, and Improving Models",
    author = "Wu, Tongshuang  and
      Ribeiro, Marco Tulio  and
      Heer, Jeffrey  and
      Weld, Daniel",
    editor = "Zong, Chengqing  and
      Xia, Fei  and
      Li, Wenjie  and
      Navigli, Roberto",
    booktitle = "Proceedings of the 59th Annual Meeting of the Association for Computational Linguistics and the 11th International Joint Conference on Natural Language Processing (Volume 1: Long Papers)",
    month = aug,
    year = "2021",
    address = "Online",
    publisher = "Association for Computational Linguistics",
    url = "https://aclanthology.org/2021.acl-long.523/",
    doi = "10.18653/v1/2021.acl-long.523",
    pages = "6707--6723"
}

@inproceedings{llm-as-jusge-withmt,
author = {Zheng, Lianmin and Chiang, Wei-Lin and Sheng, Ying and Zhuang, Siyuan and Wu, Zhanghao and Zhuang, Yonghao and Lin, Zi and Li, Zhuohan and Li, Dacheng and Xing, Eric P. and Zhang, Hao and Gonzalez, Joseph E. and Stoica, Ion},
title = {Judging {LLM}-as-a-judge with {MT}-bench and Chatbot Arena},
year = {2023},
publisher = {Curran Associates Inc.},
address = {Red Hook, NY, USA},
booktitle = {Proceedings of the 37th International Conference on Neural Information Processing Systems},
articleno = {2020},
numpages = {29},
location = {New Orleans, LA, USA},
series = {NIPS '23}
}
